\documentclass{article}

\usepackage{microtype}
\usepackage{graphicx}
\usepackage{subfig}
\usepackage{booktabs} %

\usepackage{include/titletoc}
\usepackage{amsmath}
\usepackage{amsthm}
\usepackage{amssymb}
\usepackage{natbib}
\usepackage{hyperref}

\usepackage{subfig}
\usepackage{multirow}
\usepackage{array}
\usepackage{multicol}
\usepackage{tabularx}

\usepackage{hyperref}

\usepackage[utf8]{inputenc}
\usepackage{booktabs} %
\usepackage{hyperref}

\usepackage{bbm} 
\usepackage{bm}
\usepackage{verbatim}
\usepackage{float}
\usepackage{color,soul}
\usepackage{enumitem}
\usepackage{mathtools}
\usepackage{hhline}
\usepackage[title]{appendix}
\usepackage[nameinlink]{cleveref} %
\usepackage[font=small,labelfont=bf,
   justification=justified,
   format=plain]{caption}
\usepackage{tikz}
\usepackage{wrapfig,booktabs}
\usepackage{xspace}
\usepackage{cancel}
\usepackage{sidecap}

\graphicspath{ {../figures/} }

\DeclarePairedDelimiterX{\infdivx}[2]{(}{)}{%
  #1\;\delimsize\|\;#2%
}

\newcommand{\gp}{\textsc{gp}\xspace}

\newcommand{\elbo}{\textsc{elbo}\xspace}

\newcommand{\kld}{KL divergence\xspace}

\crefname{appsec}{appendix}{appendices}
\Crefname{appsec}{Appendix}{Appendices}

\newtheorem{innercustomproposition}{Proposition}
\newenvironment{customproposition}[1]
  {\renewcommand\theinnercustomproposition{#1}\innercustomproposition}
  {\endinnercustomproposition}

\usetikzlibrary{shapes.geometric, arrows, bayesnet, calc, positioning}

\newcommand{\bX}{\mathbf X}
\newcommand{\by}{\mathbf y}
\newcommand{\bx}{\mathbf x}

\newcommand{\calF}{\mathcal{F}}

\newcommand{\calN}{\mathcal{N}}

\newcommand{\calL}{\mathcal{L}}

\newcommand{\R}{\mathbb{R}}

\newcommand{\bmu}{{\boldsymbol{\mu}}}
\newcommand{\btheta}{{\boldsymbol{\theta}}}

\newcommand{\bSigma}{\boldsymbol\Sigma}

\newcommand{\closer}[3]{{\kern-#1ex{#2}\kern-#3ex}}

\mathchardef\mhyphen="2D

\DeclareMathOperator{\E}{\mathbb{E}}

\usepackage{algorithm}
\usepackage{algorithmic}
\usepackage{colortbl}
\usepackage{tcolorbox}

\definecolor{azure}{rgb}{0.0, 0.5, 1.0}
\definecolor{airforceblue}{rgb}{0.36, 0.54, 0.66}
\definecolor{darkgreen}{rgb}{0.0, 0.2, 0.13}

\newcommand{\pms}[1]{\ensuremath{{\scriptstyle\pm #1}}}

\newcommand\defines{\doteq}

\newcommand{\frcl}{\textsc{frcl}\xspace}
\newcommand{\fromp}{\textsc{fromp}\xspace}
\newcommand{\vcl}{\textsc{vcl}\xspace}

\newcommand{\sfsvi}{\textsc{s-fsvi}\xspace}
\newcommand{\ewc}{\textsc{ewc}\xspace}
\newcommand{\si}{\textsc{si}\xspace}

\newcommand{\vargp}{\textsc{var-gp}\xspace}
\newcommand{\mnist}{\textsc{mnist}\xspace}

\newcommand{\smnist}{\textsc{s-mnist}\xspace}
\newcommand{\pmnist}{\textsc{p-mnist}\xspace}
\newcommand{\sfmnist}{\textsc{s-fmnist}\xspace}
\newcommand{\cifar}{\textsc{cifar}\xspace}

\newcommand{\vbar}{\,|\,}

\newcommand{\jac}{\mathcal{J}}
\newcommand{\calX}{\mathcal{X}}

\newcommand{\calD}{\mathcal{D}}
\newcommand{\calB}{\mathcal{B}}
\newcommand{\calQ}{\mathcal{Q}}
\newcommand{\calC}{\mathcal{C}}

\newcommand{\DD}{\mathbb{D}}

\newcommand{\bepsilon}{\boldsymbol{\epsilon}}

\newcommand{\bTheta}{\boldsymbol{\Theta}}

\newcommand{\qtilde}{\smash{\tilde{q}}}
\newcommand{\ptilde}{\smash{\tilde{p}}}
\newcommand{\flin}{\smash{\tilde{f}}}

\newcommand{\NA}{---}

\newtheorem{proposition}{Proposition}

\usepackage[algo2e]{algorithm2e}

\usepackage[nohyperref,accepted]{include/icml2022}

\icmltitlerunning{
    Continual Learning via Sequential Function-Space Variational Inference
}

\begin{document}

\raggedbottom

\twocolumn[
\icmltitle{
    Continual Learning via Sequential Function-Space Variational Inference
}

\icmlsetsymbol{equal}{*}

\begin{icmlauthorlist}
\icmlauthor{Tim G. J. Rudner}{oxford}
\icmlauthor{Freddie Bickford Smith}{oxford}
\icmlauthor{Qixuan Feng}{oxford}
\icmlauthor{Yee Whye Teh}{oxford}
\icmlauthor{Yarin Gal}{oxford}
\end{icmlauthorlist}

\icmlaffiliation{oxford}{University of Oxford, Oxford, UK}

\icmlcorrespondingauthor{Tim G. J. Rudner}{tim.rudner@cs.ox.ac.uk}

\vskip 0.3in
]

\printAffiliationsAndNotice{}  %

\begin{abstract}
Sequential Bayesian inference over predictive functions is a natural framework for continual learning from streams of data. However, applying it to neural networks has proved challenging in practice. Addressing the drawbacks of existing techniques, we propose an optimization objective derived by formulating continual learning as sequential function-space variational inference. In contrast to existing methods that regularize neural network parameters directly, this objective allows parameters to vary widely during training, enabling better adaptation to new tasks. Compared to objectives that directly regularize neural network predictions, the proposed objective allows for more flexible variational distributions and more effective regularization. We demonstrate that, across a range of task sequences, neural networks trained via sequential function-space variational inference achieve better predictive accuracy than networks trained with related methods while depending less on maintaining a set of representative points from previous tasks.
\end{abstract}

\vspace*{-10pt}
\section{Introduction}
\label{sec:intro}
Continual learning promises to enable applications of machine learning to settings with resource constraints, privacy concerns, or non-stationary data distributions.
However, continual learning in deep neural networks remains a challenge. 
While progress has been made to mitigate ``forgetting'' of previously learned abilities, existing objective-based approaches to continual learning still fall short.

A popular family of objectives penalizes changes in parameters from one task to another~\citep{ahn19,aljundi18,Chaudhry2018RiemannianWF,Kirkpatrick2017OvercomingCF,Lee2017OvercomingCF,Liu2018RotateYN,Loo2020GeneralizedVC,Nguyen2018VariationalCL,Park2019ContinualLB,Ritter2018OSL,schwarz2018progress,Swaroop2019ImprovingAU,Yin2020SOLACL,Yin2020OptimizationAG,zenke17}.
However, explicitly regularizing parameters in this way may be ineffective, since parameters are only a proxy for a neural network's predictive function.
For example, predictive functions defined by overparameterized neural networks may be obtained with several different parameter configurations, and small changes in a network’s parameters may cause large changes in its predictions.

An alternative approach that addresses this shortcoming is to regularize the \emph{predictive function} directly~\citep{benjamin2018measuring,bui2017streaming,buzzega2020dark,Jung2018LessforgetfulLF,kapoor2021variational,Kim2018KeepAL,Li2018LearningWF,morenomunoz2019continual,Pan2020ContinualDL,titsias2020functional}.
Existing function-space regularization methods represent the state of the art among objective-based approaches to continual learning~\citep{kapoor2021variational,Pan2020ContinualDL,titsias2020functional}.
Yet, as we demonstrate, these methods still leave room for improvement.
For example, ``functional regularization of the memorable past'' (\fromp; \citealp{Pan2020ContinualDL}) uses a Laplace approximation and as such does not directly optimize variance parameters, while ``functional regularization for continual learning'' (\frcl; \citealp{titsias2020functional}) is constrained to linear models.

\begin{figure*}[t]
    \centering
    \includegraphics[width=0.8\textwidth]{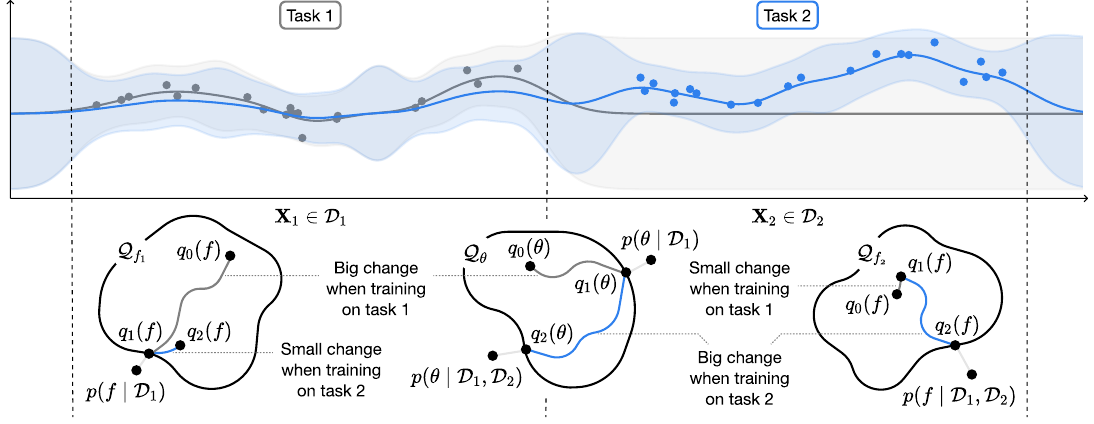}
    \caption{
        Schematic of how sequential function-space variational inference (\sfsvi) allows a Bayesian neural network to learn new tasks while maintaining previously learned abilities.
        \textit{(Top: predictive distributions.)}
        On task 1, the model fits dataset $\calD_{1}$ by updating an initial distribution over parameters $q_{0}(\btheta)$ to a variational posterior $q_{1}(\btheta)$, which in turn induces a distribution over functions $q_{1}(f)$.
        On task 2, the variational objective encourages the posterior distribution over functions to match $q_{1}(f)$ on a small set of data points from task 1 while also fitting dataset $\calD_{2}$.
        The mean and two standard deviations of the distributions over functions learned on task 1 and task 2 are shown in grey and blue, respectively.
        \textit{(Bottom: learning trajectories.)}
        On task 1, the distribution over functions changes by a large amount for inputs $\bX_1$ (left) but by a small amount for inputs $\bX_2$ (right).
        On task 2, the reverse is true.
        On both tasks, the change in the distribution over parameters (center) is decoupled from the changes in the distribution over functions (left, right).
    }
    \label{fig:illustration_fsvi_cl}
\end{figure*}

To address these limitations, we frame continual learning as sequential function-space variational inference (\sfsvi) and adapt the variational objective proposed by~\citet{Rudner2021fsvi} to the continual-learning setting.
The resulting variational optimization objective has three key advantages over existing alternatives.
First, it is expressed purely in terms of distributions over predictive functions, which allows greater flexibility than with parameter-space regularization methods~(\Cref{fig:illustration_fsvi_cl}).
Second, unlike \fromp, it allows direct optimization of variational variance parameters.
Third, unlike \frcl, it can be applied to fully-stochastic neural networks---not just to Bayesian linear models.

We demonstrate that \sfsvi outperforms existing objective-based continual learning methods---in some cases by a significant margin---on a wide range of task sequences, including single-head split \mnist, multi-head split \cifar, and multi-head sequential Omniglot.
We further present empirical results that showcase the usefulness of learned variational variance parameters and demonstrate that \sfsvi is less reliant on careful selection of datapoints that summarize past tasks than other methods.\vspace*{-10pt}

\section{Background}
\label{sec:background}

\subsection{Continual Learning as Bayesian Inference}
\label{sec:cl_bayes}
Consider a sequence of tasks indexed by $t \in \{1, \ldots, T\}$.
Each task involves making predictions on a supervised dataset $\mathcal{D}_t = (\bX_t, \by_t)$.
Continual learning is the problem of inferring a distribution over predictive functions that fits the whole collection of datasets $\{\mathcal{D}_1, \ldots, \mathcal{D}_T\}$ as well as possible given access to only a single full dataset at a time.

Sequential Bayesian inference over predictive functions $f$ provides a natural framework for this.
Assuming we have a prior $p(f)$, the posterior distribution over $f$ at task 1 is
\begin{align}
    p(f \vbar \mathcal{D}_1)
    =
    p(\mathcal{D}_1 \vbar f) p(f) / p(\mathcal{D}_1) .
\end{align}
For subsequent tasks $t$, the posterior can be expressed as
\begin{align}
\label{eq:function_posterior}
    p(f \vbar \mathcal{D}_1, \ldots,  \mathcal{D}_t) \propto p(\mathcal{D}_t \vbar f) p(f \vbar \mathcal{D}_1, \ldots,  \mathcal{D}_{t-1}) ,
\end{align}
where the posterior after task $t-1$ is treated as the prior for task $t$.
Given the intractibility of computing this posterior exactly, we need to use approximate inference.

\subsection{Function-Space Variational Inference}
\label{sec:fsvi}
Given a dataset $\mathcal{D}=(\bX,\by)$, a prior $p(f)$ and a variational family $\calQ_{f}$, function-space variational inference \citep{burt2021understanding,matthews2016kld,Rudner2021fsvi,sun2019functional} consists of finding the variational distribution $q(f) \in \calQ_{f}$ that maximizes
\begin{align}
\label{eq:elbo_function_space}
\mathbb{E}_{q(f)}[\log p(\by \vbar f(\bX))]
- \DD_\textrm{KL}(q(f) \,\|\, p(f)) .
\end{align}
This variational optimization problem presents a trade-off between fitting the data and matching a prior over functions.
To address the fact that the \kld between distributions over functions is not in general tractable, prior works have developed estimation procedures that allow turning~\Cref{eq:elbo_function_space} into an objective function that can be used in practice~\citep{Rudner2021fsvi,sun2019functional}.

\section{Continual Learning via Sequential Function-Space Variational Inference}
\label{sec:continual_fsvi}

The ideas presented in~\Cref{sec:background} provide a starting point for our method.
To approximate the posterior in~\Cref{eq:function_posterior} at task $t$, we would like to find a variational distribution $q_t(f) \in \calQ_{f}$ that minimizes
\begin{align}
    \DD_\textrm{KL}(q_t(f) \,\|\, p_t(f | \calD_{1}, ..., \calD_{t})),
\end{align}
which can equivalently be expressed as maximizing
\begin{align*}
    \mathbb{E}_{q_t(f)}[\log p(\by_t \vbar f(\bX_t))]
    \hspace*{-2pt}-\hspace*{-2pt} \DD_\textrm{KL}(q_t(f) \,\|\, p_t(f | \calD_{1}, ..., \calD_{t-1})) .
\end{align*}
Since we do not have access to $p_t(f | \calD_{1}, ..., \calD_{t-1})$, we simplify the inference problem to maximizing the variational objective
\begin{align}
\label{eq:elbo_function_space_t}
    \mathbb{E}_{q_t(f)}[\log p(\by_t \vbar f(\bX_t))]
    - \DD_\textrm{KL}(q_t(f) \,\|\, p_t(f)),
\end{align}
where for \mbox{$t=1$} we assume some prior $p_1(f)$ and for \mbox{$t>1$} the prior is given by the variational posterior distribution over functions inferred on the previous task. That is,
$$
p_t(f) \defines q_{t-1}(f) .
$$

While this objective is in general intractable for distributions over functions induced by neural networks with stochastic parameters,~\citet{Rudner2021fsvi} proposed an approximation that makes this objective amenable to gradient-based optimization and scalable to large neural networks.
To perform sequential function-space variational inference, we adapt the estimation procedure proposed by~\citet{Rudner2021fsvi} to the continual-learning setting:

\begin{proposition}[Sequential Function-Space Variational Inference (\sfsvi); adapted from~\citet{Rudner2021fsvi}]
\label{prop:continual_fsvi}
Let $D_{t}$ be the number of model output dimensions for $t$ tasks, let \mbox{$f^{\text{\emph{}}} : \calX \times \R^P \rightarrow \R^{D_{t}}$} be a mapping defined by a neural network architecture, let $\bTheta \in \R^P$ be a multivariate random vector of network parameters, and let $q_{t}(\btheta) \defines \calN(\bmu_{t}, \bSigma_{t})$ and $q_{t-1}(\btheta) \defines \calN(\bmu_{t-1}, \bSigma_{t-1})$ be variational distributions over $\bTheta$.
Additionally, let $\bX_{\calC}$ denote a set of context points, and let \mbox{$\bar{\bX}_{t} \subseteq \{ \bX_{t} \cup \bX_{\calC} \}$}.
Under a diagonal approximation of the prior and variational posterior covariance functions across output dimensions, the objective in \Cref{eq:elbo_function_space_t} can be approximated by
\begin{align}
\begin{split}
\label{eq:variational_objective_paper}
    & \calF(q_{t}, q_{t-1}, \bX_{\calC}, \bX_{t}, \by_{t})
    \\
    &
    \defines \E_{q_{t}(\btheta)}[\log p(\by_{t} \vbar f(\bX_{t} ; \btheta) ) ]
    \\
    & ~~~
    -
    \sum_{k=1}^{D_{t}}
    \frac{1}{2} \bigg(\hspace*{-1pt} \log \frac{|[\mathbf{K}^{p_t}]_{k}|}{|[\mathbf{K}^{q_t}]_{k}|} -\hspace*{-1pt} \frac{|\bar{\bX}_{t}|}{D_{t}}\hspace*{-3pt} + \text{\emph{Tr}}([\mathbf{K}^{p_t}]_{k}^{-1}[\mathbf{K}^{q_t}]_{k}) \hspace*{-1pt}
    \\
    & ~~~ ~~~
    + \Delta(\bar{\bX}_{t}; \bmu_{t}, \bmu_{t-1})^\top [\mathbf{K}^{p_t}]_{k}^{-1} \Delta(\bar{\bX}_{t}; \bmu_{t}, \bmu_{t-1}) \bigg),
    \end{split}
\end{align}
where
    \begin{align}
    \Delta(\bar{\bX}_{t}; \bmu_{t}, \bmu_{t-1}) \defines [f(\bar{\bX}_{t}; \bmu_{t})]_{k} - [f(\bar{\bX}_{t}; \bmu_{t-1})]_{k}
\end{align}
and
\begin{align}
    \mathbf{K}^{p_t} &\defines
        \jac(\bar{\bX}_{t}, \bmu_{t-1}) \bSigma_{t-1} \jac(\bar{\bX}_{t}, \bmu_{t-1})^\top
    \\
    \mathbf{K}^{q_t} &\defines
        \jac(\bar{\bX}_{t}, \bmu_{t}) \bSigma_t \jac(\bar{\bX}_{t}, \bmu_{t})^\top
    ,
\end{align}\\[-10pt]
are covariance matrix estimates constructed from Jacobians $\jac(\cdot, \mathbf{m}) \defines \frac{\partial f(\cdot \,; \bTheta)}{\partial \bTheta}|_{\bTheta = \mathbf{m}}\,$ with \mbox{$\mathbf{m} = \{\bmu_{t}, \bmu_{t-1} \}$}.
\end{proposition}
\begin{proof}
See~\Cref{appsec:proofs}.
\end{proof}

``Functional regularization for continual learning'' (\frcl;~\citealp{titsias2020functional}) and ``functional regularization of the memorable past'' (\fromp;~\citealp{Pan2020ContinualDL}) use objectives conceptually similar to the objective in~\Cref{eq:elbo_function_space_t} and mathematically similar to the objective in~\Cref{eq:variational_objective_paper}.
To highlight the differences between the \sfsvi objective above and \fromp and \frcl, respectively, we make the relationship between these two methods and \sfsvi precise in the following two propositions.

\newpage

\begin{proposition}[Relationship between \fromp and \sfsvi]
\label{prop:fromp_fsvi}
With the \sfsvi objective $\calF$ defined as in \Cref{eq:variational_objective_paper}, let $\bar{\bX}_{t} = \bX_{\calC}$.
Then, up to a multiplicative constant, the \fromp objective corresponds to the \sfsvi objective with the prior covariance given by a Laplace approximation about $\bmu_{t-1}$ and the variational distribution given by a Dirac delta distribution $q_t^{\fromp}(\btheta) \defines \delta(\btheta - \bmu_{t})$.
Denoting the prior covariance under a Laplace approximation about $\bmu_{t-1}$ by $\hat{\bSigma}_0(\bmu_{t-1})$ so that \mbox{$q_{t-1}^{\fromp}(\btheta) \defines \calN(\bmu_{t-1}, \hat{\bSigma}_0(\bmu_{t-1}))$}, the \fromp objective can be expressed as
\begin{align*}
    \begin{split}
    & \calL^{\fromp}(q_{t}^{\fromp}, q_{t-1}^{\fromp}, \bX_{\calC}, \bX_{t}, \by_{t})
    \\
    & ~~~
    = \calF(q_{t}^{\fromp}, q_{t-1}^{\fromp}, \bX_{\calC}, \bX_{t}, \by_{t}) - \mathcal{V} ,
    \end{split}
\end{align*}
where
\begin{align*}
    \mathcal{V}
    \defines - \frac{1}{2} \sum_{k}
    \left(
        \log \frac{{[\bar{\mathbf{K}}^{\hat{p}_t}]_{k}}}{{[\bar{\mathbf{K}}^{q_t}]_{k}}} + \frac{[\bar{\mathbf{K}}^{q_t}]_{k}}{[\bar{\mathbf{K}}^{\hat{p}_t}]_{k}} - 1
    \right) ,
\end{align*}\\[-5pt]
with $\bar{\mathbf{K}}$ denoting a covariance matrix under a block-diagonalization without inter-task dependence, and
\begin{align*}
    \bar{\mathbf{K}}^{\hat{p}_t}
    \hspace*{-2pt}\defines\hspace*{-2pt} \text{\emph{block-diag}}\hspace*{-2pt}\left( \jac(\bar{\bX}_{t}, \bmu_{t-1}) \hat{\bSigma}_{0}(\bmu_{t-1}) \jac(\bar{\bX}_{t}, \bmu_{t-1})^\top \hspace*{-2pt}\right)\hspace*{-2pt} .
\end{align*}
\end{proposition}
\begin{proof}
See~\Cref{appsec:proofs}.
\end{proof}
\Cref{prop:fromp_fsvi} shows that the \fromp objective nearly corresponds to the \sfsvi objective but is missing the term in the \sfsvi objective (denoted by $\mathcal{V}$ above) that encourages learning variational variance parameters that accurately reflect the variance of the prior.
This insight reflects a shortcoming of the \fromp objective.
Unlike in the \sfsvi objective which allows optimization over $\bSigma$, the \fromp objective is restricted to covariance estimates given by the Laplace approximation.

The \frcl objective can be related to the \sfsvi objective in a similar way:
\begin{proposition}[Relationship between \frcl and \sfsvi]
\label{prop:frcl_fsvi}
With the \sfsvi objective $\calF$ defined as in \Cref{eq:variational_objective_paper}, let $\bar{\bX}_{t} = \bX_{\calC}$, and let \mbox{$f^{\text{\emph{LM}}}(\cdot \,; \bTheta) \defines \Phi_{\psi}(\cdot) \bTheta$} be a Bayesian linear model, where $\Phi_{\psi}(\cdot)$ is a deterministic feature map parameterized by $\psi$.
Then the \frcl objective corresponds to the \sfsvi objective for the model $f^{\text{\emph{LM}}}(\cdot \,; \bTheta)$ plus an additional weight-space \kld penalty.
That is, for \mbox{$p_t(\btheta) \defines \calN(\mathbf{0}, \mathbf{I}$}, and $q_t(\btheta) \defines \calN(\bmu_{t}, \bSigma_t)$,
\begin{align}
    \begin{split}
    & \calL^{\text{\frcl}}(q_{t}^{\frcl}, q_{t-1}^{\frcl}, \bX_{\calC}, \bX_{t}, \by_{t})
    \\
    & ~~~
    = \calF(q_{t}^{\frcl}, q_{t-1}^{\frcl}, \bX_{\calC}, \bX_{t}, \by_{t}) + \DD_{\textrm{\emph{KL}}}(q_t(\btheta) \,\|\, p_t(\btheta)) .
    \end{split}
\end{align}
\end{proposition}
\begin{proof}
See~\Cref{appsec:proofs}.
\end{proof}
\Cref{prop:frcl_fsvi} highlights that the \frcl objective is restricted to Bayesian linear models and does not regularize the deterministic parameters in the feature map as effectively as if they were variational parameters.

\begin{figure*}[t!]
    \centering
    \hspace*{-10pt}
    \subfloat[Task 1]{
        \label{fig:toy_2d_1}
        \includegraphics[height=2.4cm]{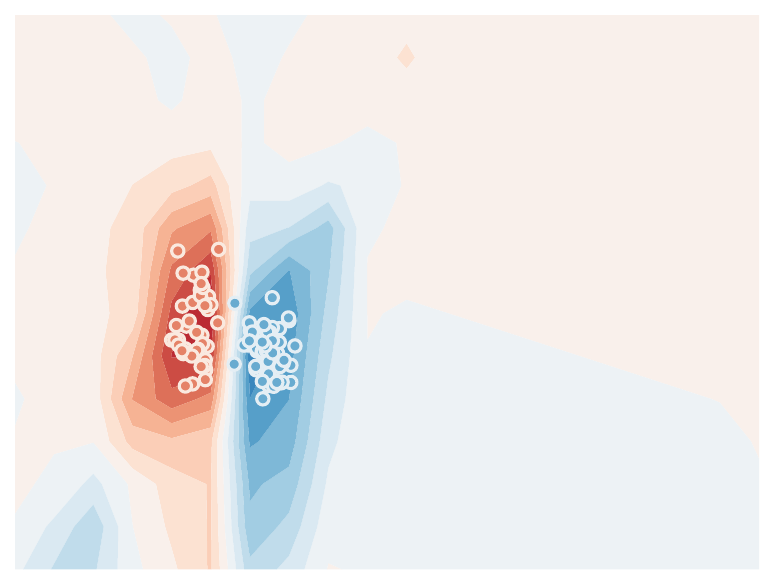}
    }
    \hspace*{-12pt}
    \subfloat[Task 2]{
        \label{fig:toy_2d_2}
        \includegraphics[height=2.4cm]{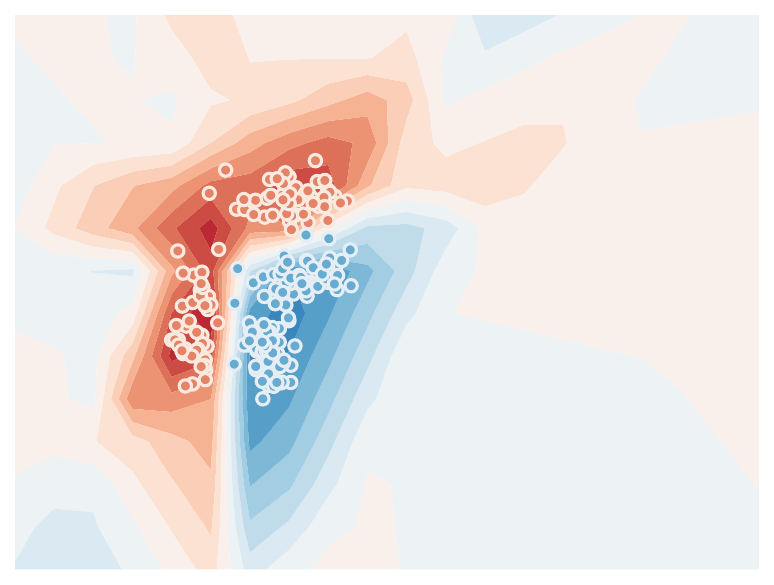}
    }
    \hspace*{-12pt}
    \subfloat[Task 3]{
        \label{fig:toy_2d_3}
        \includegraphics[height=2.4cm]{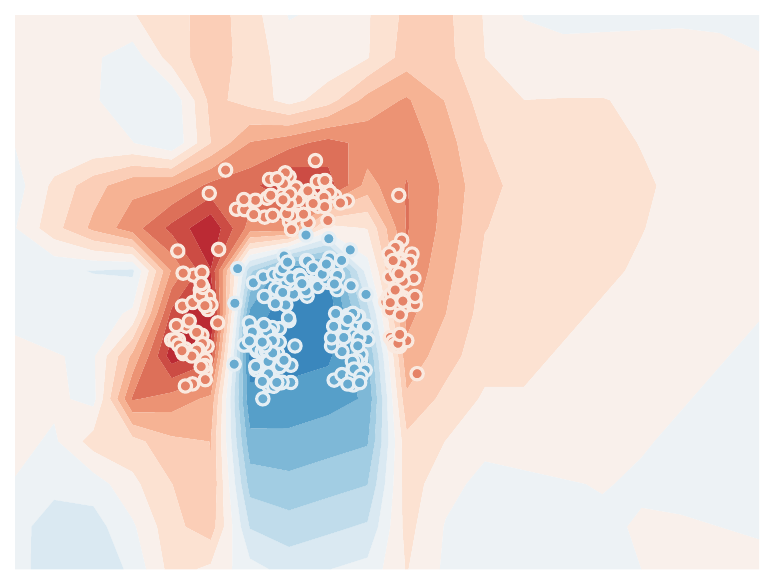}
    }
    \hspace*{-12pt}
    \subfloat[Task 4]{
        \label{fig:toy_2d_4}
        \includegraphics[height=2.4cm]{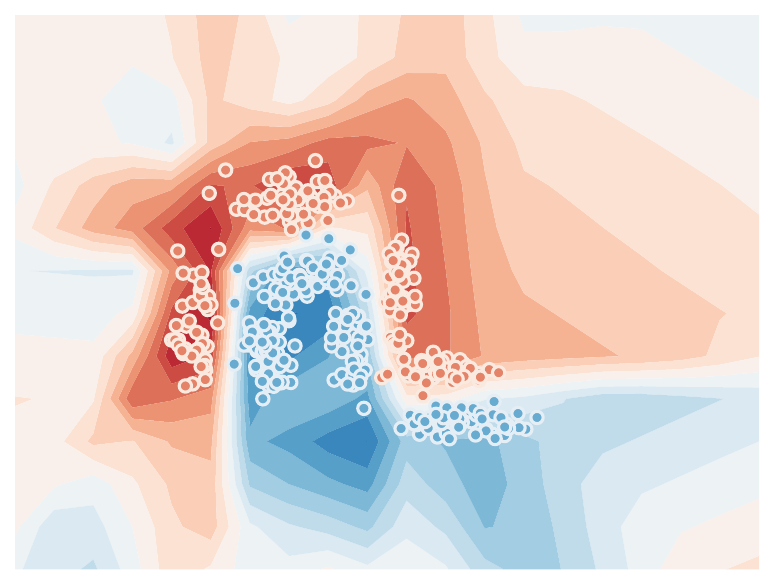}
    }
    \hspace*{-12pt}
    \subfloat[Task 5]{
        \label{fig:toy_2d_5}
        \includegraphics[height=2.4cm]{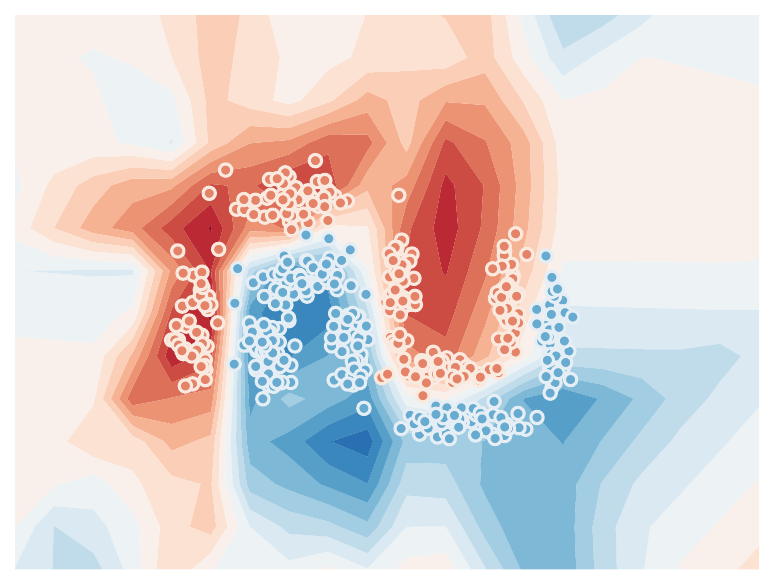}
    }
    \hspace*{-5pt}
    \subfloat{
        \includegraphics[height=2.4cm]{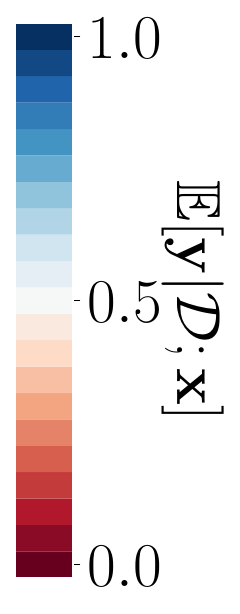}
    }
    \caption{
         A practical demonstration of sequential function-space variational inference (\sfsvi) on a sequence of five binary-classification tasks with 2D inputs.
        The neural network infers a decision boundary between the two classes while maintaining high predictive uncertainty away from the data.
        The experimental setup is described in detail in~\Cref{appsec:experiment_details}.
    }
    \label{fig:2d_toy}
    \vspace*{-5pt}
\end{figure*}

\subsection{Simplified Sequential Function-Space VI}

For ease of computation and to ensure scalability to large neural networks, we consider mean-field distributions $q^{\text{MF}}_t(\btheta)$ for all tasks, diagonalize the covariance matrix estimates $\mathbf{K}^{p_t}$ and $\mathbf{K}^{q_t}$ across input points in $\bar{\bX}_{t}$, and let $(\bX_{\calB}, \by_{\calB}) \subset \mathcal{D}_t$ be a mini-batch from the current dataset.
This way, we obtain the simplified variational objective
\begin{align}
\begin{split}
\label{eq:variational_objective_diag_paper}
    & \tilde{\calF}(q^{\text{MF}}_{t}, q^{\text{MF}}_{t-1}, \bX_{\calC}, \bX_{\calB}, \by_{\calB})
    \\
    &
    = \frac{1}{S} \sum_{i=1}^{S} \log p(\by_{\calB} \vbar f(\bX_{\calB} ; h(\bmu_{t}, \bSigma_t, \bepsilon^{(i)})) )
    \\
    & ~~~
    -
    \sum_{j=1}^{|\bar{\bX}|} \sum_{k=1}^{D_{t}}
    \frac{1}{2} \bigg( \log \frac{[\mathbf{K}^{p_t}]_{j,k}}{[\mathbf{K}^{q_t}]_{j,k}} + \frac{[\mathbf{K}^{q_t}]_{j,k}}{[\mathbf{K}^{p_t}]_{j,k}} - 1
    \\
    & \quad \quad \quad
    + \frac{\left([f(\bar{\bX}_{t}; \bmu_{t})]_{j,k} - [f(\bar{\bX}_{t}; \bmu_{t-1})]_{j,k} \right)^2}{[\mathbf{K}^{p_t}]_{j,k}} \bigg),
    \end{split}
\end{align}
where $h(\bmu_{t}, \bSigma_t, \bepsilon^{(i)}) \defines \bmu_{t} + \bSigma_t \odot \bepsilon^{(i)}$ is a reparameterization of $\bTheta \in \mathbb{R}^P$ with $\bepsilon^{(i)} \sim \calN(\mathbf{0}, \mathbf{I}_{P})$, $S$ is the number of Monte Carlo samples, $D_{t}$ is as defined before, and
\begin{align}
    \mathbf{K}^{p_t} &\defines
    \text{{diag}}
    \left(
        \jac(\bar{\bX}_{t}, \bmu_{t-1}) \bSigma_{t-1} \jac(\bar{\bX}_{t}, \bmu_{t-1})^\top
    \right)
    \\
    \mathbf{K}^{q_t} &\defines
    \text{{diag}}
    \left(
        \jac(\bar{\bX}_{t}, \bmu_{t}) \bSigma_t \jac(\bar{\bX}_{t}, \bmu_{t})^\top
    \right) .
\end{align}
This simplified objective does not require matrix inversion, and the time and space complexity for gradient estimation and prediction scale linearly in the number of context points $\bar{\bX}_{t}$ and network parameters.
The context set $\bX_{\calC}$ can be constructed from coresets containing representative points from previous tasks.

We provide an empirical comparison of the simplified \sfsvi, \fromp, and \frcl objectives in~\Cref{sec:experiments} to assess the extent to which the differences described above affect continual learning.

\setlength{\tabcolsep}{14pt}
\begin{table*}[t!]
\vspace*{-10pt}
    \caption{
        Predictive accuracies of a selection of objective-based methods for continual learning.
        Results are reported for three task sequences: split \mnist (\smnist), split Fashion \mnist (\sfmnist) and permuted \mnist (\pmnist).
        In some cases, a multi-head setup (MH) is used; in others, a single-head setup (SH).
        Best results for identical network architectures are printed in boldface (exception: \textsc{var-gp} uses a non-parametric model).
        Best overall results are highlighted in gray.
        Each numerical entry denotes the mean accuracy across tasks at the end of training.
        Where possible, this accuracy is based on experiments repeated with different random seeds (10 repeats for \sfsvi), with both the mean value and standard error reported.
        All methods use the same architecture and coreset size unless indicated otherwise.
        See \Cref{appsec:experiment_details} for more experimental details.
        $^{1}$Accuracies computed using the best coreset-selection method (either random or $k$-center).
        $^{2}$Uses random coreset selection.
        $^{3}$Requires a multi-head setup with task identifiers, including for permuted \mnist.
        This requirement explains the missing \frcl result for \smnist (SH).
        $^{4}$Uses a larger MLP architecture (see~\Cref{tab:hyperparameter} in appendix).%
        $^{5}$Evaluates the \kld at points sampled from the empirical data distribution of the current task.
        $^{6}$Uses one sample per class as a coreset. %
    }
    \centering
    \small
    \vspace*{2pt}
    \label{tab:results}
    \begin{tabular}{l c c c c}
        \toprule
        Method & \multicolumn{1}{c}{\smnist (MH)} & \multicolumn{1}{c}{\sfmnist (MH)} & \multicolumn{1}{c}{\pmnist (SH)} & \multicolumn{1}{c}{\smnist (SH)} \\
        \midrule
        \ewc~\citep{Kirkpatrick2017OvercomingCF} & 63.10\% & \NA & 84.00\% & \NA \\
        \si~\citep{zenke17} & 98.90\% & \NA & 86.00\% & \NA \\
        \vcl~\citep{Nguyen2018VariationalCL}$^{1}$ &  98.40\% & 98.60\%\pms{0.04}  & 93.00\% & 32.11\%\pms{1.16}
        \\
        $~~~$\vcl (no coreset)  & 97.00\% & 89.60\%\pms{1.75} & 87.50\%\pms{0.61} & 17.74\%\pms{1.20}
        \\
        \frcl~\citep{titsias2020functional}$^{3}$ & 97.80\%\pms{0.22} & 97.28\%\pms{0.17}  & 94.30\%\pms{0.06} & \NA
        \\
        \fromp~\citep{Pan2020ContinualDL} & {99.00\%}\pms{0.04} & 99.00\%\pms{0.03} & 94.90\%\pms{0.04} & 35.29\%\pms{0.52} \\
        \vargp~\citep{kapoor2021variational} & \NA & \NA & \textbf{97.20\%}\pms{0.08} & 90.57\%\pms{1.06} \\
        \sfsvi (\textbf{ours})$^{2}$ & \textbf{99.54\%}\pms{0.04} & \textbf{99.19\%}\pms{0.02} & 95.76\%\pms{0.02} & \textbf{92.87\%}\pms{0.14}
        \\
        \midrule
        \sfsvi Ablation Study: &  &  &  & 
        \\
        $~~~$\sfsvi (larger networks)$^{4}$ & \cellcolor[gray]{0.9}\textbf{99.76\%}\pms{0.00} & 99.16\%\pms{0.03} & \cellcolor[gray]{0.9}\textbf{97.50\%}\pms{0.01} & \cellcolor[gray]{0.9}\textbf{93.38\%}\pms{0.10}
        \\
        $~~~$\sfsvi (no coreset)$^{5}$ & 99.62\%\pms{0.02} & \cellcolor[gray]{0.9}\textbf{99.54\%}\pms{0.01} & 84.06\%\pms{0.46} & 20.15\%\pms{0.52} \\
        $~~~$\sfsvi (minimal coreset)$^{6}$ & \NA & \NA & 89.59\%\pms{0.30} & 51.44\%\pms{1.22} \\
        \bottomrule
    \end{tabular}
    \vspace*{-10pt}
\end{table*}

\section{Related Work}
\label{sec:related_work}
There are three main (partially overlapping) categories of methods for continual learning in a deep neural network.
Objective-based approaches modify the objective function used to train the neural network.
Replay-based approaches summarize past tasks using either stored data or freshly generated synthetic data.
Architecture-based approaches change the neural network's structure from one task to another.
For extensive reviews, see~\citet{DeLange2021ACL} and~\citet{Parisi2019ContinualLL}.
As sequential function-space variational inference (\sfsvi) centers around a new training objective, we focus on objective-based approaches in this review.
(Like the methods reviewed below, \sfsvi does incorporate a form of replay in that it uses context points, but the primary interest is the training objective.)

For a neural network to retain abilities it has previously learned, its predictions on data associated with past tasks must not change significantly from one task to another.
One way of achieving this is to include in the training objective a form of function-space regularization to discourage important changes in the network's predictions or internal representations.
``Learning without forgetting''~\citep{Li2018LearningWF} uses a modified cross-entropy loss that penalizes the difference between the predictions of the current network on the current task data and the predictions of the previous network on the current task data.
``Less-forgetful learning''~\citep{Jung2018LessforgetfulLF} employs the same method but uses squared Euclidean distance rather than the modified cross-entropy loss and applies it to the penultimate-layer representations rather than the network's predictions.
``Keep and learn''~\citep{Kim2018KeepAL} also uses internal representations as a basis for regularization.
The method subsequently proposed by~\citet{benjamin2018measuring} involves comparing the current network with all previous versions of the network and on data from all past tasks instead of with only the most recent network on data from the current task.
Each pair of networks is compared by computing the Euclidean distance between the networks' predictions.
``Dark experience replay''~\citep{buzzega2020dark} extends this method to work in a setting where task boundaries are not clearly defined.

While these approaches mitigate forgetting, they do not explicitly account for predictive uncertainty, which is an issue if the neural network is a poor fit to the data.
This deficiency is addressed by probabilistic approaches to function-space regularization, which encourage a network's predictions to agree with a prior distribution over functions rather than with a single function.
``Functional regularization for continual learning'' (\frcl;~\citealp{titsias2020functional}) considers a network whose final layer is a Bayesian linear model.
Based on the duality between parameter space and function space, the \frcl objective includes the \kld between predictive distributions at a selection of input points.
This encourages similarity between the network's current predictive distribution and the distributions from past tasks.
\frcl is theoretically appealing, building on a well-understood method for stochastic variational inference using inducing points, but is only applicable to Bayesian linear models.
In contrast, ``functional regularization of the memorable past'' (\fromp; ~\citealp{Pan2020ContinualDL}) maintains a posterior distribution over all the parameters of a neural network.
While \fromp achieves state-of-the-art performance on several continual-learning task sequences, it relies on a change in the underlying probabilistic model and uses a surrogate objective for optimization, which divorces it from function-space variational objectives.
As we show, this results in suboptimal performance compared to sequential function-space variational inference, which maintains a stronger link to the underlying Bayesian approximation.

Although our focus is on methods for training deep neural networks, for completeness, we also note methods based on Gaussian processes ({\gp}s).
Incremental variational sparse \gp regression~\citep{cheng2016incremental}, streaming sparse {\gp}s~\citep{bui2017streaming} and online sparse multi-output \gp regression~\citep{Yang2019Online} built on the work of~\citet{csato2002sparse} and~\citet{csato2002gaussian}, and are effective approaches to continual learning for regression tasks.
Continual multi-task {\gp}s~\citep{morenomunoz2019continual} extend to multi-output settings with non-Gaussian likelihoods.
The success of variational autoregressive {\gp}s~(\vargp; ~\citealp{kapoor2021variational}) on continual learning for task sequences with image inputs gives reason for inclusion where relevant in~\Cref{sec:experiments}.
However, we note that \vargp scales poorly with the number of tasks: the time complexity for inference is cubic in the number of context points and hence in the number of tasks, which may limit its applicability to task sequences like sequential Omniglot.
In contrast, the time complexity of \sfsvi is linear in the number of context points.

Also distinct from but related to our method are a number of objective-based approaches to continual learning that directly regularize the parameters of a neural network.
We briefly discuss these approaches in~\Cref{appsec:related_work}.

\begin{figure*}[t!]
\vspace*{-7pt}
    \centering
    \hspace*{-15pt}
    \subfloat[\smnist (MH)]{%
        \label{fig:coreset_heuristic_smnist_mh}
        \includegraphics[height=3.2cm]{
            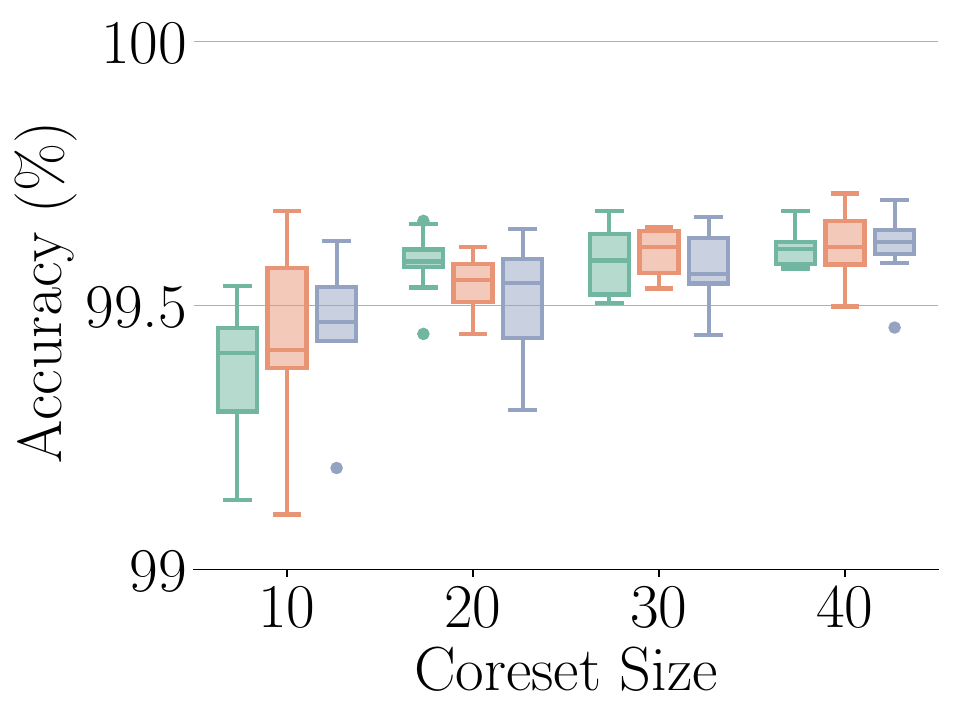}%
    }
    \subfloat[\sfmnist (MH)]{%
        \label{fig:coreset_heuristic_sfmnist_mh}
        \includegraphics[height=3.2cm]{
            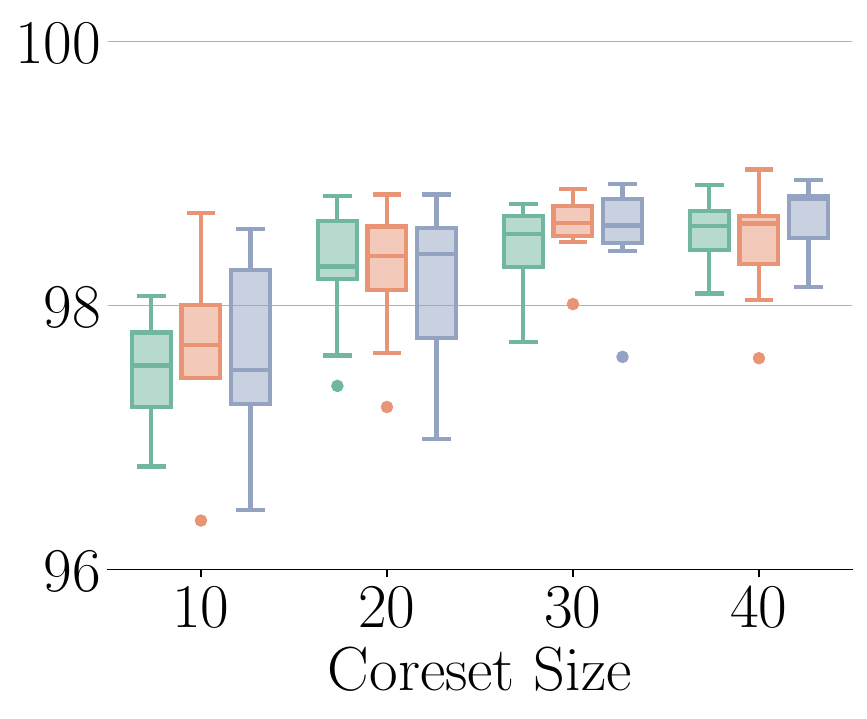}%
    }
    \subfloat[\pmnist (SH)]{%
        \label{fig:coreset_heuristic_pmnist}
        \includegraphics[height=3.2cm]{
            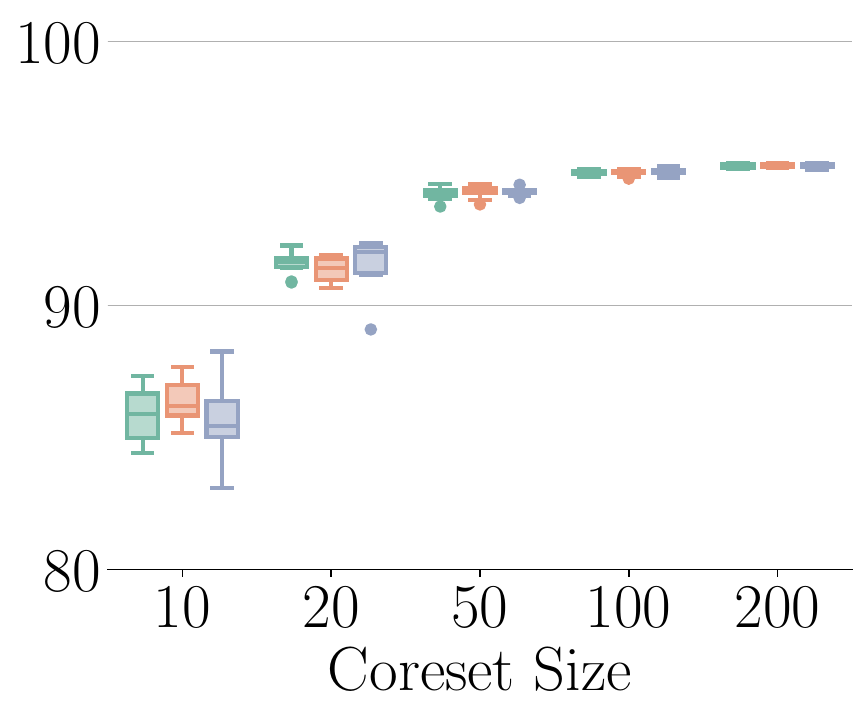}%
    }
    \subfloat[\smnist (SH)]{%
        \label{fig:coreset_heuristic_smnist_sh}
        \includegraphics[height=3.2cm]{
            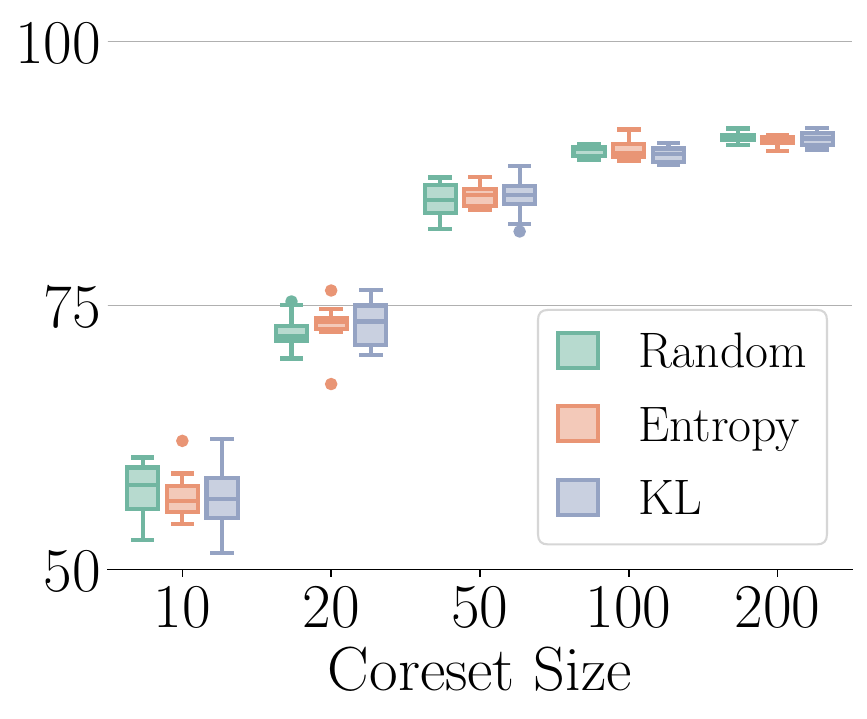}%
    }
    \caption{
        Effect of the coreset size and coreset-selection method on the predictive accuracy of \sfsvi.
        Three coreset-selection methods are presented: sampling data points with uniform probability; sampling with probability proportional to model's predictive entropy; and sampling with probability proportional to the \kld between the posterior predictive distribution and the prior predictive distribution.
        Ten inducing points are used in each case.
        No coreset-selection method consistently yields higher accuracy.
    }
    \label{fig:coreset_heuristic}
    \vspace*{-5pt}
\end{figure*}

\section{Empirical Evaluation}
\label{sec:experiments}
After visualizing how \sfsvi works in practice (\Cref{sec:uncertainty}), we compare \sfsvi's performance with that of existing objective-based methods for continual learning (\Cref{sec:mnist,sec:omniglot,sec:cifar}).
For a comprehensive comparison, we evaluate \sfsvi on a range of task sequences used in related work.
Aiming to use as strong baselines as possible, we report results taken directly from the literature in most cases (and mention when we do not).
Reporting baselines in this way leaves gaps in our comparison: for each existing technique, results are available for only a subset of the task sequences we consider here (e.g.,~\citet{Pan2020ContinualDL} report results for split \cifar but not sequential Omniglot, while~\citet{titsias2020functional} do the reverse).

Our evaluation pays attention to two factors important in the assessment of continual-learning methods: the use of task identifiers when making predictions, and the use of a coresets of data points to summarize past tasks~\citep{farquhar_towards_2018}.
To provide some commentary on the first of these factors, we run an experiment that compares the performance of a single-head neural network (which does not use task identifiers) to that of a multi-head neural network (which uses task identifiers).
Regarding the second factor, we explore how performance changes when the coreset size changes or a context set unrelated to previous tasks is used.

Details about the experimental setups (e.g., optimization routines and hyperparameter searches) can be found in~\Cref{appsec:experiment_details}.
Our code can be accessed at:
\begin{tcolorbox}[colback=blue!5,colframe=blue!75!black]
\begin{center}
    \href{https://timrudner.com/sfsvi-code}{\texttt{https://timrudner.com/sfsvi-code}}.
\end{center}
\end{tcolorbox}

\begin{figure*}[t!]
\vspace*{-7pt}
    \centering
    \hspace*{-20pt}
    \subfloat[Split \cifar Accuracies After Training on All Tasks]{
        \label{fig:cifar_fromp}
        \includegraphics[height=4.8cm]{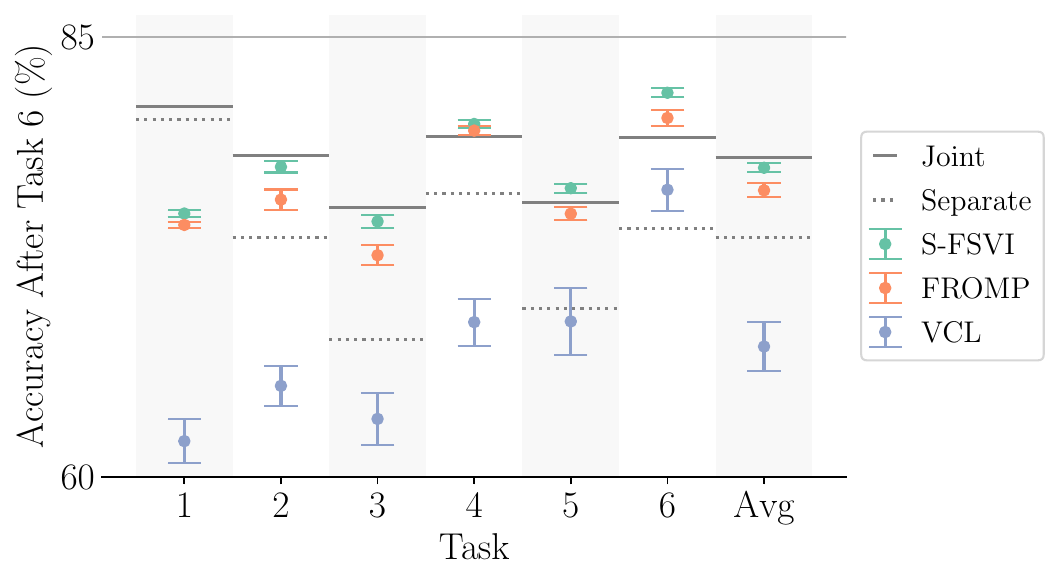}
    }
    \hspace*{15pt}
    \subfloat[\sfsvi Accuracy as a Function of Coreset Size]{
        \label{fig:coreset_heuristic_cifar}
        \includegraphics[height=4.8cm]{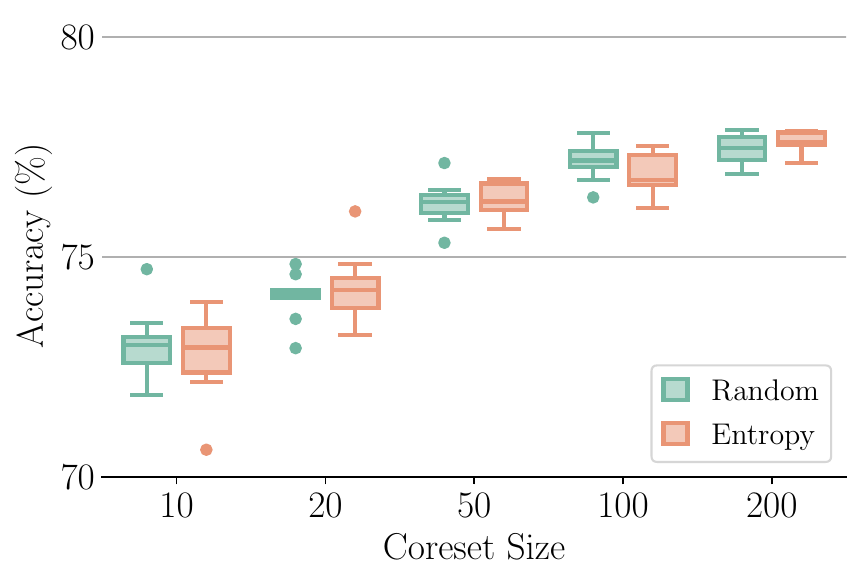}
    }
    \label{fig:cifar}
    \vspace*{-3pt}
    \caption{
        Predictive accuracies of \sfsvi and related methods on split \cifar.
        (a) Per-task and average accuracy after training on six tasks.
        The result of ``joint'' baseline is obtained using a model trained on data from all tasks at the same time.
        The accuracy at task $t$ for the ``separate'' baseline is the accuracy of an independent model trained only on task $t$.
        We use the best performing method for each baseline: \fromp for ``joint'', \sfsvi for ``separate''.
        (b) Average accuracy after training on six tasks with different coreset sizes.
        ``Random'' coreset selection denotes uniform sampling from the training set.
        ``Entropy'' coreset selection denotes sampling from the training set with probability proportional to the entropy of the model's posterior predictive distribution.
    }
    \vspace*{-10pt}
\end{figure*}

\subsection{Illustrative Example}
\label{sec:uncertainty}
To provide intuition for how \sfsvi allows learning on new tasks while maintaining previously acquired abilities, we apply it to a task sequence based on easy-to-visualize synthetic 2D data, originally proposed by~\citet{Pan2020ContinualDL}.
In this task sequence, each data point belongs to one of two classes, and more data points are revealed as the task sequence progresses.
The data-generating process is assumed to reveal data from mostly non-overlapping subsets of the input space.
The continual-learning problem is then to infer the decision boundary around data points revealed up to and including the current task without forgetting the decision boundary inferred on previous tasks.
We use a single-head neural network.

In~\Cref{fig:2d_toy}, we plot the model's posterior predictive distribution after training on each of five tasks.
After training on task 1, the model has low predictive uncertainty close to the data points and high uncertainty (class probabilities around 0.5) everywhere else (\Cref{fig:toy_2d_1}).
On task 2, \sfsvi seeks to match the distribution over functions inferred on the previous task while fitting the new set of data points.
\sfsvi achieves this and expands the area in input space where the model is confident in its predictions (\Cref{fig:toy_2d_2}).

As more tasks and data are revealed, \sfsvi allows the model to continually explore the data space and infer the decision boundary while maintaining accurate, high-confidence predictions on data points in parts of the inputs space where it was previously trained on observed data.
Finally, after training on five tasks, the model has inferred the decision boundary between the two classes, while maintaining high predictive uncertainty in parts of the input space where no data points have been observed yet (\Cref{fig:toy_2d_5}).
The model maintains high predictive uncertainty away from the data, which makes it easier to learn on new tasks.
This is unlike deterministic neural networks, which tend to make highly confident predictions in parts of the inputs space where no data has been observed, or on data points that lie outside of the distribution of the training data.

\subsection{Split (Fashion) MNIST \& Permuted MNIST}
\label{sec:mnist}

Having established some intuition for how \sfsvi works, we demonstrate how this translates to high predictive accuracy on three task sequences commonly used to evaluate continual-learning methods.
First is split \mnist (\smnist), in which each task consists of binary classification on a pair of \mnist classes (0 vs. 1, 2 vs. 3, and so on).
Second is split Fashion \mnist (\sfmnist), which has the same structure but uses data from Fashion \mnist, posing a harder problem.
Third is permuted \mnist (\pmnist), in which each task consists of ten-way classification on \mnist images whose pixels have been randomly reordered.
A multi-head setup (MH) with task identifiers provided at prediction time is the default for \smnist and \sfmnist, while a single-head setup (SH) without task identifiers is standard for \pmnist.
In addition to running the default setup for all three task sequences, we run a single-head setup for \smnist.

With a standard configuration, \sfsvi outperforms all existing methods based on deep neural networks by a statistically significant margin on all task sequences (\Cref{tab:results}).
As noted in \Cref{sec:related_work}, \vargp's conceptual connection to our method warrants its inclusion in our comparison.
\vargp performs better than our standard configuration of \sfsvi on permuted \mnist, but this advantage disappears once a larger neural network is used with \sfsvi.
Moreover, \vargp is unlikely to scale well to more challenging task sequences, such as those in \Cref{sec:omniglot,sec:cifar}.

\setlength{\tabcolsep}{18.5pt}
\begin{table}[b!]
\vspace*{-12pt}
    \caption{
        Predictive accuracies of \sfsvi and related methods on sequential Omniglot.
        For \sfsvi and \frcl, the coreset consists of two data points per class.
        All baseline results are from~\citet{titsias2020functional}.
        For all methods, the mean and standard deviation over five random task permutations are reported.
        $^{1}$\citet{Li2018LearningWF}.
        $^{2}$\citet{schwarz2018progress}.
        $^{3}$\citet{schwarz2018progress}.
        $^{4}$Coreset selected using \frcl's ``trace'' method.
        $^{5}$Details in \Cref{appsec:experiment_details}.
    }
    \centering
    \small
    \label{tab:omniglot}
    \vspace*{2pt}
    \begin{tabular}{l c}
        \toprule
        Method & \multicolumn{1}{c}{Test Accuracy} \\
        \midrule
        Learning Without Forgetting$^{1}$ & 62.06\%\pms{2.0} \\
        \ewc & 67.32\%\pms{4.7} \\
        Online \textsc{ewc}$^{2}$ & 69.99\%\pms{3.2} \\
        Progress \& Compress$^{3}$ & 70.32\%\pms{3.3} \\
        \textsc{frcl}$^{4}$ & 81.47\%\pms{1.6} \\
        \sfsvi (\textbf{ours})$^{5}$ & \cellcolor[gray]{0.9}\hspace*{2pt}\textbf{83.29\%}\pms{1.2}\hspace*{4pt} \\
        \bottomrule
    \end{tabular}
    \vspace*{-10pt}
\end{table}

\vspace*{-4pt}
\subsection{Sequential Omniglot}
\label{sec:omniglot}

\begin{figure}[h!]
\vspace*{-10pt}
    \centering
    \includegraphics[width=\linewidth,trim={0 15pt 0 0}]{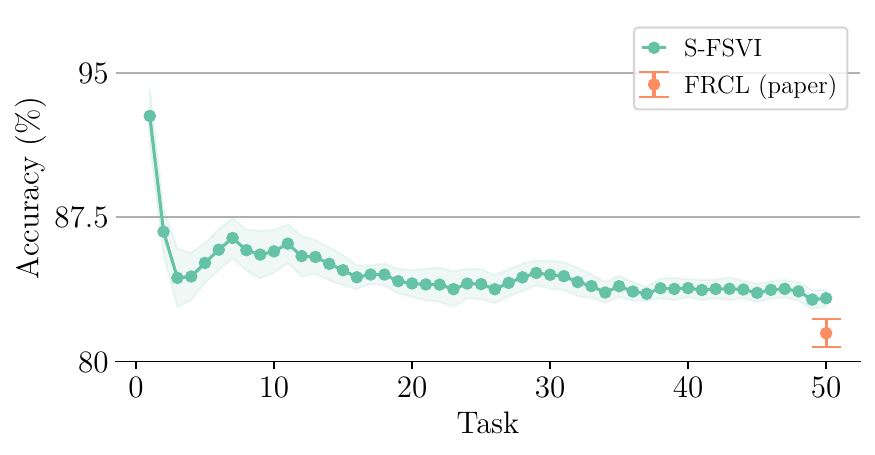}
    \caption{
        Predictive accuracies of \sfsvi and \frcl on sequential Omniglot.
        For \sfsvi, the accuracy shown at task $t$ is the mean accuracy across all tasks up to that point (mean $\pm$ one standard error as computed across five permutations of the task order).
        We were unable to reproduce the result reported in~\citet{titsias2020functional} using the authors' code.
        However, we compare against the result from the paper (only the accuracy at task 50 is reported) here to provide a strong baseline.
    }
    \label{fig:omniglot}
    \vspace*{-8pt}
\end{figure}

Sequential Omniglot~\citep{lake2015human,schwarz2018progress} provides a more challenging task sequence than those considered in~\Cref{sec:mnist}.
It consists of 50 classification tasks, where the number of classes varies between the tasks (details in~\Cref{appsec:experiment_details}).
We find that \sfsvi produces better predictive accuracy than all available baselines, including \frcl, by a statistically significant margin (\Cref{tab:omniglot}).
To illustrate the stability of \sfsvi across long task sequences, we plot its mean accuracy over 50 tasks in~\Cref{fig:omniglot}.

\begin{figure*}[t!]
\vspace*{-8pt}
    \begin{minipage}[l]{0.48\textwidth}
    \hspace*{-8pt}\subfloat[\smnist (MH)]{
        \label{fig:vcl_smnist}
        \includegraphics[height=3.28cm]{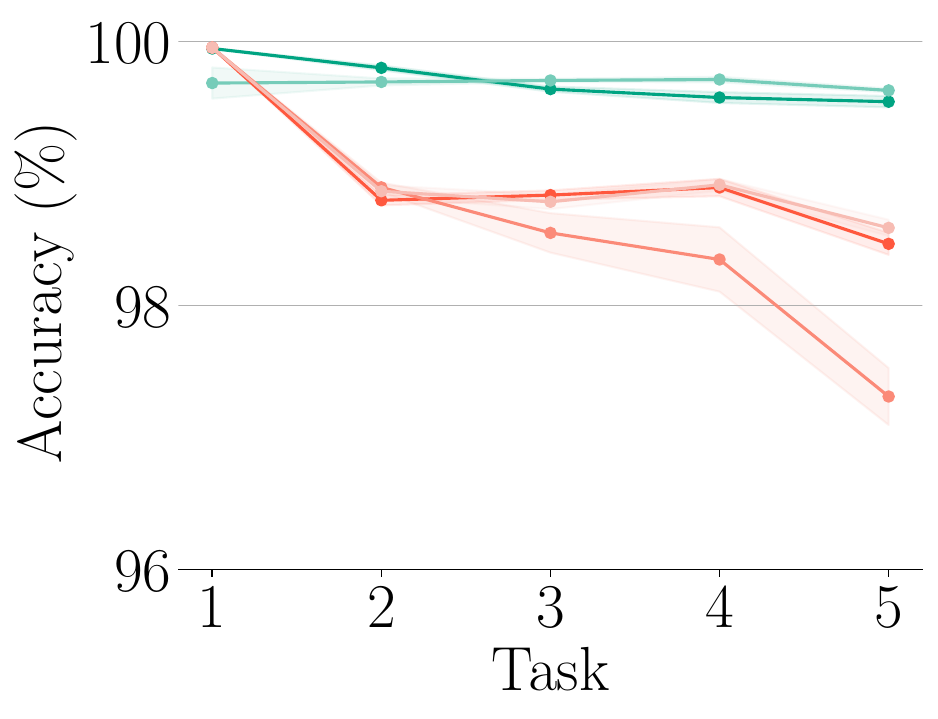}}
    \hspace*{-8pt}\subfloat[\pmnist (SH)]{
        \label{fig:vcl_pmnist}
        \includegraphics[height=3.28cm]{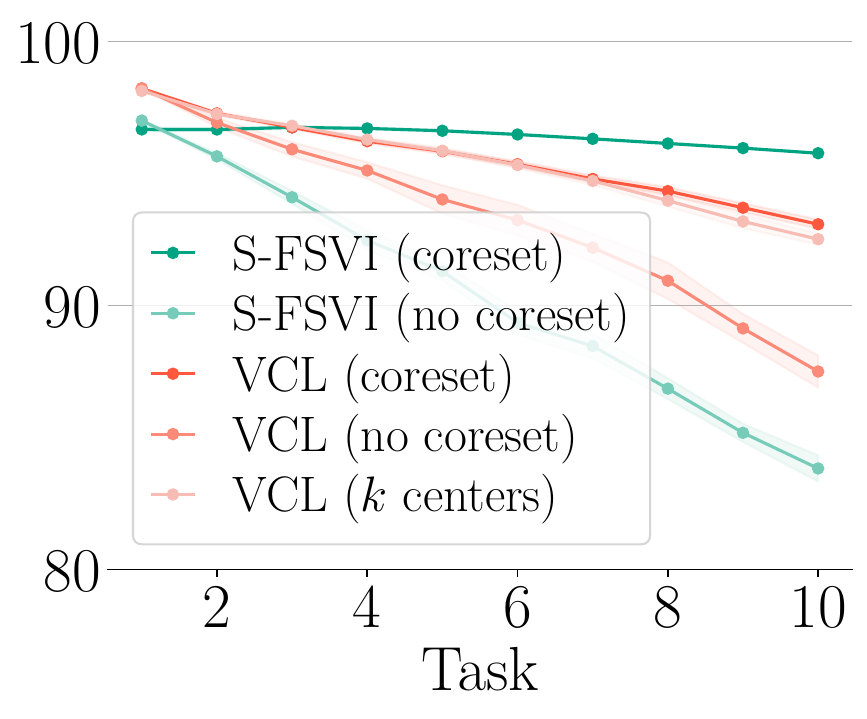}}
    \caption{
        Predictive accuracies of \sfsvi and parameter-space variational inference (\vcl) on split \mnist and permuted \mnist.
        The accuracy shown at task $t$ is the mean accuracy across all tasks up to that point (mean $\pm$ one standard error as computed across ten repetitions of the experiment).
        With a coreset, \sfsvi outperforms \vcl on both task sequences.
        Without a coreset, \sfsvi performs poorly on permuted \mnist.
    }
    \label{fig:vcl}
    \end{minipage}
    \hspace*{15pt}
    \begin{minipage}[r]{0.48\textwidth}
    \vspace*{2pt}
    \hspace*{-8pt}\subfloat[\smnist (MH)]{
        \label{fig:coreset_none_smnist}
        \includegraphics[height=3.2cm,trim={0 10pt 0 10pt},clip]{
            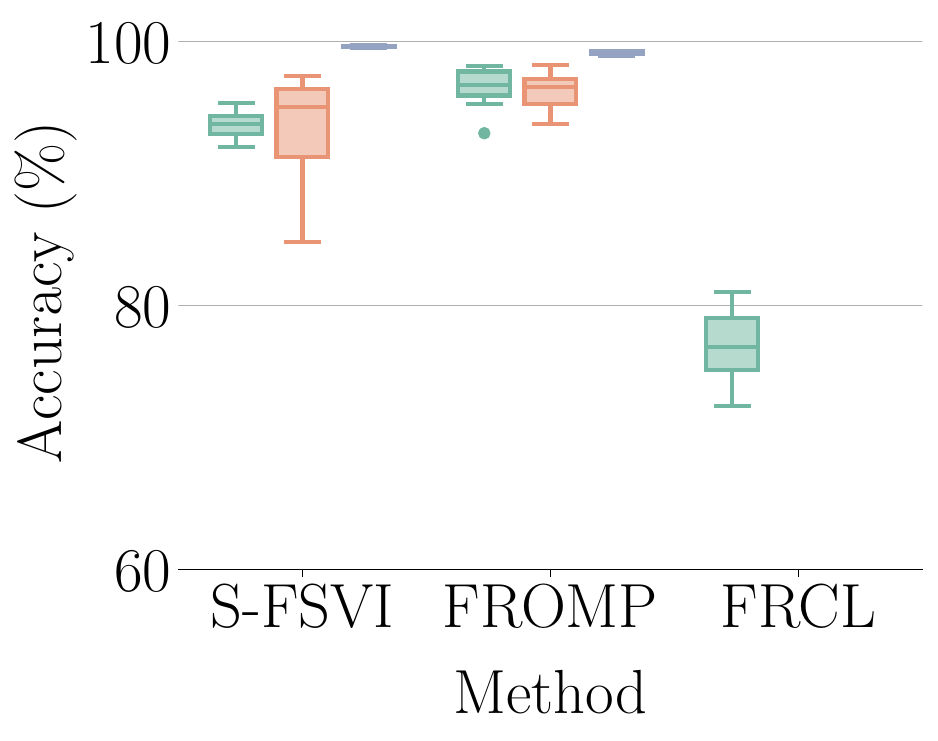}
    }
    \hspace*{-8pt}\subfloat[\sfmnist (MH)]{
        \label{fig:coreset_none_sfmnist}
        \includegraphics[height=3.2cm,trim={0 10pt 0 10pt},clip]{
            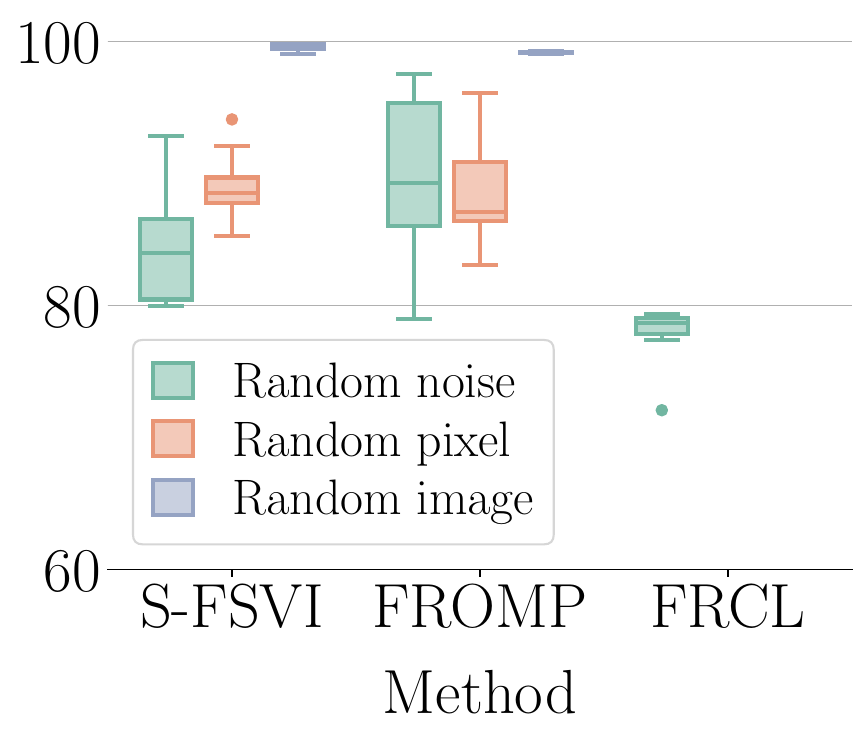}
    }
    \caption{
        Predictive accuracies of \sfsvi, \fromp and \frcl on multi-head split (Fashion) \mnist without using coresets.
        Inducing inputs for evaluating the \kld are sampled according to three different sampling schemes derived from the current task's empirical data distribution (see \Cref{appsec:experiment_details} for details).
        Using \sfsvi with images sampled from the current task's training set significantly outperforms all other methods.
    }
    \label{fig:coreset_none}
    \end{minipage}
    \vspace*{-14pt}
\end{figure*}

\vspace*{-8pt}
\subsection{Split CIFAR}
\label{sec:cifar}
Moving beyond classification tasks on grayscale images, we evaluate \sfsvi on split \cifar~\citep{Pan2020ContinualDL,zenke17}.
This uses the full \cifar-10 dataset for the first task, followed by five ten-way classification tasks drawn from \cifar-100.
Our results show \sfsvi achieving higher accuracy on all tasks than \fromp and \vcl after learning all six tasks (\Cref{fig:cifar_fromp}).
Notably, on each task except the first, \sfsvi performs close to or better than two baselines: a model trained only on that task, and a model trained on all tasks jointly.
The latter is a particularly strong baseline, because all data is available during training.

As in related work~\citep{lopez2017episodic,Pan2020ContinualDL}, we compute the forward transfer (FT) and backward transfer (BT) for \sfsvi on split \cifar.
FT captures by how much the accuracy on the current tasks increases as the number of past tasks increases; BT captures by how much the accuracy on the previous tasks increases as more tasks are observed (see \Cref{appsec:forward_backward_transfer} for mathematical definitions).
As well as having the best overall accuracy, \sfsvi significantly outperforms all baselines in terms of FT and has BT comparable to \ewc and \fromp (\Cref{tab:transfer}).

\setlength{\tabcolsep}{9.0pt}
\begin{table}[h!]
\vspace*{-8pt}
    \caption{
        Forward transfer (FT) and backward transfer (BT) of \sfsvi and related methods on split \cifar.
        All baseline results are from~\citet{Pan2020ContinualDL}.
        For all methods, the mean and standard error over five repeated experiments are reported.
        $^{1}$Details in \Cref{appsec:experiment_details}.
    }
    \centering
    \small
    \label{tab:transfer}
    \vspace*{2pt}
    \begin{tabular}{l c c c}
        \toprule
        Method & \multicolumn{1}{c}{Test Accuracy} & \multicolumn{1}{c}{FT} & \multicolumn{1}{c}{BT} \\
        \midrule
        \ewc & 71.6\%\pms{0.4} & 0.2\pms{0.4} & \textbf{-2.3}\pms{0.6} \\
        \vcl & 67.4\%\pms{0.6} & 1.8\pms{1.4} & -9.2\pms{0.8} \\
        \fromp& 76.2\%\pms{0.2} & 6.1\pms{0.3} & \textbf{-2.6}\pms{0.4} \\
        \sfsvi (ours)$^{1}$ & \cellcolor[gray]{0.9}\textbf{77.6\%}\pms{0.2}\hspace*{4pt} & \cellcolor[gray]{0.9}\textbf{7.3}\pms{0.2} & \cellcolor[gray]{0.9}\textbf{-2.5}\pms{0.2} \\
        \bottomrule
    \end{tabular}
    \vspace*{-7pt}
\end{table}

\subsection{\mbox{Function- vs. Parameter-Space Inference}}

To demonstrate the importance of performing inference in function space, we compare how the accuracies of \sfsvi and \vcl evolve from one task to another on split \mnist and permuted \mnist~(\Cref{fig:vcl}).
We find that \sfsvi consistently outperforms \vcl whose predictive performance steadily degrades suggesting that function-space inference may be more effective than parameter-space inference at transferring prior knowledge from one task to another, and that this may offset the information loss in the \kld between distributions over functions compared to the \kld between distributions over parameters.
\subsection{Coreset Size and Selection}
Similar to existing methods such as \fromp and \frcl, \sfsvi includes in the training objective a function-space regularization term that encourages matching the prior distribution over functions at a set of context points.
Typically, this requires keeping a representative coreset of data points from each task, from which a context set can be constructed.

\sfsvi offers two benefits with respect to coresets.
First, it is insensitive to which points get included in the coresets. 
Whereas existing methods often require expensive procedures to select important data points from previous tasks, \Cref{fig:coreset_heuristic,fig:coreset_heuristic_cifar} show that \sfsvi achieves strong performance while only using randomly selected coresets.
Second, \sfsvi does not require large coresets to perform well.
On permuted \mnist, \sfsvi achieves better predictive accuracy than \ewc and \si even if the coreset used for \sfsvi consists of only a single data point per class (\Cref{tab:results}).
On the single-head version of split \mnist, a minimal coreset (one point per class, or two points per task) allows \sfsvi to outperform \vcl and \fromp, both with coresets of 40 points per task (\Cref{tab:results}).
In some multi-head settings, \sfsvi achieves state-of-the-art predictive accuracies with randomly-generated noise coresets (\Cref{tab:results} and \Cref{fig:coreset_none}).

\section{Conclusion}
\label{sec:conclusion}

We presented sequential function-space variational inference (\sfsvi), a method for continual learning in deep neural networks.
We showed that \sfsvi improves on the predictive performance of existing objective-based continual learning methods---often by a significant margin---including on task sequences with high-dimensional inputs (split \cifar) and large numbers of tasks (sequential Omniglot).
Lastly, we demonstrated that---unlike existing function-space regularization methods---\sfsvi does not rely on careful coreset selection and, in multi-head settings, can achieve state-of-the-art performance even without coresets collected on previous tasks.
We hope that this work will lead to future research into further improving function-space objectives for continual learning.\vspace*{-10pt}

\section*{Acknowledgements}

Tim G. J. Rudner and Freddie Bickford Smith are funded by the Engineering and Physical Sciences Research Council (EPSRC).
Tim G. J. Rudner
is also funded by the Rhodes Trust and by a Qualcomm Innovation Fellowship.
We gratefully acknowledge donations of computing resources by the Alan Turing Institute.

\bibliography{references}
\bibliographystyle{apalike}

\clearpage

\appendix
\begin{appendices}

\crefalias{section}{appsec}
\crefalias{subsection}{appsec}
\crefalias{subsubsection}{appsec}

\setcounter{equation}{0}
\renewcommand{\theequation}{\thesection.\arabic{equation}}

\onecolumn

\section*{\LARGE Supplementary Material}
\label{sec:appendix}

\vspace*{20pt}
\section*{Table of Contents}

\begin{enumerate}[label=(\Alph*)]
    \item[]
    \textbf{\Cref{appsec:proofs}:} Proofs
    \item[]
    \textbf{\Cref{appsec:further_empirical_results}:} Further Empirical Results
    \item[]
    \textbf{\Cref{appsec:experiment_details}:} Experimental Details
    \item[]
    \textbf{\Cref{appsec:related_work}:} Further Related Work
\end{enumerate}
\vspace*{5pt}

\section{Proofs}
\label{appsec:proofs}

\subsection{
    Variational Objective
}
\label{appsec:proof_variational_objective}

\begin{customproposition}{1}[Sequential Function-Space Variational Inference (\sfsvi); adapted from~\citep{Rudner2021fsvi}]
Let $D_{t}$ be the number of model output dimensions for $t$ tasks, let \mbox{$f^{\text{\emph{}}} : \calX \times \R^P \rightarrow \R^{D_{t}}$} be a mapping defined by a neural network architecture, let $\bTheta \in \R^P$ be a multivariate random vector of network parameters, and let $q_{t}(\btheta) \defines \calN(\bmu_{t}, \bSigma_{t})$ and $q_{t-1}(\btheta) \defines \calN(\bmu_{t-1}, \bSigma_{t-1})$ be variational distributions over $\bTheta$.
Additionally, let $\bX_{\calC}$ denote a sample of context points, and let \mbox{$\bar{\bX}_{t} \subseteq \{ \bX_{t} \cup \bX_{\calC} \}$}.
Under a diagonal approximation of the prior and variational posterior covariance functions across output dimensions, the objective in \Cref{eq:elbo_function_space_t} can be approximated by
\begin{align}
\begin{split}
\label{eq-app:variational_objective_paper}
    & \calF(q_{t}, q_{t-1}, \bX_{\calC}, \bX_{t}, \by_{t})
    \\
    &
    \defines \E_{q_{t}(\btheta)}[\log p(\by_{t} \vbar f(\bX_{t} ; \btheta) ) ]
    \\
    & \qquad
    -
    \sum_{k=1}^{D_{t}}
    \frac{1}{2} \bigg(\hspace*{-1pt} \log \frac{|[\mathbf{K}^{p_t}]_{k}|}{|[\mathbf{K}^{q_t}]_{k}|} - \frac{|\bar{\bX}_{t}|}{D_{t}} + \text{\emph{Tr}}([\mathbf{K}^{p_t}]_{k}^{-1}[\mathbf{K}^{q_t}]_{k}) + \Delta(\bar{\bX}_{t} ; \bmu_{t}, \bmu_{t-1})^\top [\mathbf{K}^{p_t}]_{k}^{-1} \Delta(\bar{\bX}_{t} ; \bmu_{t}, \bmu_{t-1}) \bigg),
    \end{split}
\end{align}
where
\begin{align}
\Delta(\bar{\bX}_{t} ; \bmu_{t}, \bmu_{t-1}) \defines [f(\bar{\bX}_{t} ; \bmu_{t})]_{k} - [f(\bar{\bX}_{t} ; \bmu_{t-1})]_{k}
\end{align}
and
\begin{align}
\SwapAboveDisplaySkip
\mathbf{K}^{p_t}
\defines
    \jac(\bar{\bX}_{t} , \bmu_{t-1}) \bSigma_{t-1} \jac(\bar{\bX}_{t} , \bmu_{t-1})^\top
\quad \text{and} \quad
\mathbf{K}^{q_t}
\defines
    \jac(\bar{\bX}_{t} , \bmu_{t}) \bSigma_t \jac(\bar{\bX}_{t} , \bmu_{t})^\top
,
\end{align}
are covariance matrix estimates constructed from Jacobians $\jac(\cdot, \mathbf{m}) \defines \frac{\partial f(\cdot \,; \bTheta)}{\partial \bTheta}|_{\bTheta = \mathbf{m}}\,$ with \mbox{$\mathbf{m} = \{\bmu_{t}, \bmu_{t-1} \}$}.
\end{customproposition}
\begin{proof}
The results follows directly from the variational objective derived in~\citep{Rudner2021fsvi} when setting the prior to $p \defines q_{t-1}$ and specifying the context set to be constructed from the coreset.
\end{proof}

\subsection{
    Derivation of Correspondence to Other Function-Space Objectives
}
\label{appsec:proof_relationship}

\begin{customproposition}{2}[Relationship between \fromp and \sfsvi]
\label{app-prop:relationship_fsvi}
With the \sfsvi objective $\calF$ defined as in \Cref{eq:variational_objective_paper}, let $\bar{\bX}_{t}  = \bX_{\calC}$.
Then, up to a multiplicative constant, the \fromp objective corresponds to the \sfsvi objective with the prior covariance given by a Laplace approximation about $\bmu_{t-1}$ and the variational distribution given by a Dirac delta distribution $q_t^{\fromp}(\btheta) \defines \delta(\btheta - \bmu_{t})$.
Denoting the prior covariance under a Laplace approximation about $\bmu_{t-1}$ by $\hat{\bSigma}_0(\bmu_{t-1})$ so that \mbox{$q_{t-1}^{\fromp}(\btheta) \defines \calN(\bmu_{t-1}, \hat{\bSigma}_0(\bmu_{t-1}))$}, the \fromp objective can be expressed as
\begin{align*}
    \begin{split}
    & \calL^{\fromp}(q_{t}^{\fromp}, q_{t-1}^{\fromp}, \bX_{\calC}, \bX_{t}, \by_{t})
    =
    \calF(q_{t}^{\fromp}, q_{t-1}^{\fromp}, \bX_{\calC}, \bX_{t}, \by_{t}) - \mathcal{V} ,
    \end{split}
\end{align*}
where
\begin{align*}
\SwapAboveDisplaySkip
    \mathcal{V}
    \defines - \frac{1}{2} \sum_{k}
    \left(
        \log \frac{{[\bar{\mathbf{K}}^{\hat{p}_t}]_{k}}}{{[\bar{\mathbf{K}}^{q_t}]_{k}}} + \frac{[\bar{\mathbf{K}}^{q_t}]_{k}}{[\bar{\mathbf{K}}^{\hat{p}_t}]_{k}} - 1
    \right) ,
\end{align*}
with $\bar{\mathbf{K}}$ denoting a covariance matrix under a block-diagonalization without inter-task dependence, and
\begin{align*}
    \bar{\mathbf{K}}^{\hat{p}_t}
    \hspace*{-2pt}\defines\hspace*{-2pt} \text{\emph{block-diag}}\hspace*{-2pt}\left( \jac(\bar{\bX}_{t} , \bmu_{t-1}) \hat{\bSigma}_{0}(\bmu_{t-1}) \jac(\bar{\bX}_{t} , \bmu_{t-1})^\top \hspace*{-2pt}\right)\hspace*{-2pt} .
\end{align*}
\end{customproposition}

\begin{proof}
By Equation (8) in~\citet{Pan2020ContinualDL}, the \fromp objective function is given by
\begin{align}
\begin{split}
    &
    \calL^{\text{\fromp}}(q_{t}^{\fromp}, q_{t-1}^{\fromp}, \bX_{\calC}, \bX_{t}, \by_{t})
    \\
    &
    \defines \E_{q_{t}(\btheta)}[\log p(\by_{t} \vbar f(\bX_{t} ; \bmu_{t}) ) ]
    \hspace*{-1.5pt}+\hspace*{-1.5pt}
    \sum_{k=1}^{t-1} \frac{\tau}{2} \left([f(\bX_{\calC}; \bmu_{t})]_{k} \hspace*{-1.5pt}-\hspace*{-1.5pt} [f(\bX_{\calC}; \bmu_{t-1})]_{k} \right)^{\top} [\mathbf{K}^{\hat{p}_{t}}]^{-1}_{k} \left([f(\bX_{\calC}; \bmu_{t})]_{k} \hspace*{-1.5pt}-\hspace*{-1.5pt} [f(\bX_{\calC}; \bmu_{t-1})]_{k} \right),
\end{split}
\end{align}\\[-10pt]
with temperature parameter $\tau$.
The result follows directly from the definition of $\calF(q_{t}^{\fromp}, q_{t-1}^{\fromp}, \bX_{\calC}, \bX_{t}, \by_{t})$ and \mbox{$\tau = 1$}.
\end{proof}

\begin{customproposition}{3}[Relationship between \frcl and \sfsvi]
\label{app-prop:frcl_fsvi}
With the \sfsvi objective $\calF$ defined as in \Cref{eq:variational_objective_paper}, let $\bar{\bX}_{t}  = \bX_{\calC}$, and let \mbox{$f^{\text{\emph{LM}}}(\cdot \,; \bTheta) \defines \Phi_{\psi}(\cdot) \bTheta$} be a Bayesian linear model, where $\Phi_{\psi}(\cdot)$ is a deterministic feature map parameterized by $\psi$.
Then the \frcl objective corresponds to the \sfsvi objective for the model $f^{\text{\emph{LM}}}(\cdot \,; \bTheta)$ plus an additional weight-space \kld penalty.
That is, for \mbox{$p_t(\btheta) \defines \calN(\mathbf{0}, \mathbf{I})$}, and $q_t(\btheta) \defines \calN(\bmu_{t}, \bSigma_t)$,
\begin{align}
    \begin{split}
    & \calL^{\text{\frcl}}(q_{t}^{\frcl}, q_{t-1}^{\frcl}, \bX_{\calC}, \bX_{t}, \by_{t})
    =
    \calF(q_{t}^{\frcl}, q_{t-1}^{\frcl}, \bX_{\calC}, \bX_{t}, \by_{t}) + \DD_{\textrm{\emph{KL}}}(q_t(\btheta) \,\|\, p_t(\btheta)) .
    \end{split}
\end{align}
\end{customproposition}

\begin{proof}
By Section 2.3 in~\citet{titsias2020functional}, the \frcl objective function is given by
\begin{align}
\begin{split}
\calL^{\text{\frcl}}(\bmu_{t}, \bSigma_{t}, \bX_{\calC}, \bX_{t}, \by_{t})
    &
    \defines
    \mathbb{E}_{q_{t}(\btheta)}[\log p(\by_{t} \vbar \Phi_{\psi}(\bX_{t}) \btheta)] - \DD_{\textrm{{KL}}}(q_{t}(\btheta) \,\|\, p_{t}(\btheta))
    \\
    &\quad
    - \DD_{\textrm{KL}}(\qtilde_{t}(\flin( \bX_{\calC_{t}} ; \btheta)) \,\|\, \ptilde_{t}(\flin( \bX_{\calC_{t}} ; \btheta)) )
    - \sum_{k=1}^{t-1} \DD_{\textrm{KL}}(\perp(\qtilde_{k}(\flin( \bX_{\calC_{k}} ; \btheta))) \,\|\, \ptilde_{k}(\flin( \bX_{\calC_{k}} ; \btheta)) ) ,
\end{split}
\end{align}
with the inducing points associated with task $k$ denoted by $\bX_{\calC_{k}}$ and $\perp$ denoting the \texttt{stop-gradient} operator, whereas the \sfsvi objective for a Bayesian linear model is
\begin{align}
\begin{split}
    & \calF(q_{t}, q_{t-1}, \bX_{\calC}, \bX_{t}, \by_{t})
    \defines \E_{q_{t}(\btheta)}[\log p(\by_{t} \vbar \Phi_{\psi}(\bX_{t}) \btheta) ) ]
    \\
    & \qquad
    - \sum_{k=1}^{D_{t}}
    \frac{1}{2} \bigg(\hspace*{-1pt} \log \frac{|[\mathbf{K}^{p_t}]_{k}|}{|[\mathbf{K}^{q_t}]_{k}|} - \frac{|\bar{\bX}_{t}|}{D_{t}} + \text{{Tr}}([\mathbf{K}^{p_t}]_{k}^{-1}[\mathbf{K}^{q_t}]_{k}) + \Delta(\bar{\bX}_{t} ; \bmu_{t}, \bmu_{t-1})^\top [\mathbf{K}^{p_t}]_{k}^{-1} \Delta(\bar{\bX}_{t} ; \bmu_{t}, \bmu_{t-1}) \bigg),
\end{split}
\end{align}
with
\begin{align}
\SwapAboveDisplaySkip
    \Delta(\bar{\bX}_{t}; \bmu_{t}, \bmu_{t-1})
    \defines
    [\Phi_{\psi}(\bar{\bX}_{t}) \bmu_{t}]_{k} - [\Phi_{\psi}(\bar{\bX}_{t}) \bmu_{t-1}]_{k}
\end{align}
and
\begin{align}
\SwapAboveDisplaySkip
    \mathbf{K}^{p_t}
    \defines
    \Phi_{\psi}(\bar{\bX}_{t}) \bSigma_{t-1} \Phi_{\psi}(\bar{\bX}_{t})^\top
    \qquad
    \mathbf{K}^{q_t}
    \defines
    \Phi_{\psi}(\bar{\bX}_{t}) \bSigma_t \Phi_{\psi}(\bar{\bX}_{t})^\top
    .
\end{align}
Letting $\bX_{\calC_{k}}$ be the context points associated with task $k$ and letting $\bar{\mathbf{K}}$ denote a covariance matrix under a block-diagonalization without inter-task dependence, we define
\begin{align}
\begin{split}
    \bar{\mathbf{K}}^{p_{t}}
    \defines
    \textrm{{block-diag}}\left( \Phi_{\psi}(\bar{\bX}_{t}) \bSigma_{t-1} \Phi_{\psi}(\bar{\bX}_{t})^\top \right)
    \qquad
    \bar{\mathbf{K}}^{q_{t}}
    \defines
    \textrm{{block-diag}}\left( \Phi_{\psi}(\bar{\bX}_{t}) \bSigma_t \Phi_{\psi}(\bar{\bX}_{t})^\top \right) ,
\end{split}
\end{align}
with diagonal entries $\{ \mathbf{K}^{p_{t}}_1, ..., \mathbf{K}^{p_{t}}_{t} \}$ and $\{ \mathbf{K}^{q_{t}}_1, ..., \mathbf{K}^{q_{t}}_{t} \}$, respectively, where each $\mathbf{K}^{p_{t}}_{k}$ is computed from task-specific context points $\bX_{\calC_{k}}$.
Fixing $[\bmu_{t-1}]_{k} = \mathbf{0}$ and $[\bSigma_{t-1}]_{k} = \mathbf{I}_{M_{k}}$ for all $k \leq t$ with $M_{k} = |\bX_{\calC_{k}}|$, as in~\citet{titsias2020functional}, we then get
\begin{align}
\SwapAboveDisplaySkip
    \bar{\mathbf{K}}^{p_{t}}_{k}
    =
    \Phi_{\psi}(\bar{\bX}_{\calC_{k}}) \Phi_{\psi}(\bar{\bX}_{\calC_{k}})^\top \quad \forall k \leq t .
\end{align}
Considering $[\bmu_{t}]_{k}$ and $[\bSigma_{t}]_{k}$ as fixed for all $k \leq t-1$, as in~\citet{titsias2020functional}, using the \texttt{stop-gradient} operator $\perp$, we can write the \sfsvi objective as
\begin{align}
\begin{split}
    & \calF(q_{t}, q_{t-1}, \bX_{\calC}, \bX_{t}, \by_{t})
    \defines \E_{q_{t}(\btheta)}[\log p(\by_{t} \vbar \Phi_{\psi}(\bX_{t}) \btheta) ) ]
    \\
    & \qquad
    - \DD_{\textrm{KL}}(\qtilde_{t}(\flin( \bX_{\calC_{t}} ; \btheta)) \,\|\, \ptilde_{t}(\flin( \bX_{\calC_{t}} ; \btheta)) )
    - \sum_{k=1}^{t-1} \DD_{\textrm{KL}}(\perp(\qtilde_{k}(\flin( \bX_{\calC_{k}} ; \btheta))) \,\|\, \ptilde_{k}(\flin( \bX_{\calC_{k}} ; \btheta)) ) ,
\end{split}
\end{align}
concluding the proof.
\end{proof}

\clearpage

\section{Further Empirical Results}
\label{appsec:further_empirical_results}

\begin{figure*}[ht!]
    \centering
    \vspace*{-10pt}
    \hspace{-10pt}
    \subfloat[\smnist (MH)]{
        \label{fig:covariance_smnist}
        \includegraphics[height=3.2cm, keepaspectratio]{
            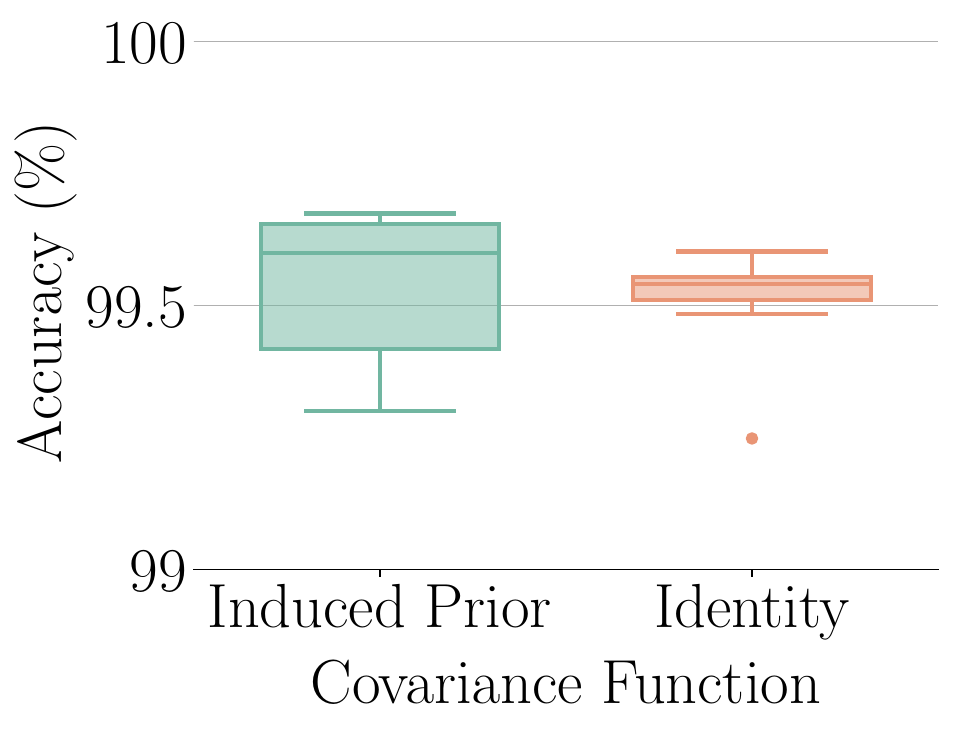}
    }
    \subfloat[\sfmnist (MH)]{
        \label{fig:covariance_sfmnist}
        \includegraphics[height=3.2cm, keepaspectratio]{
            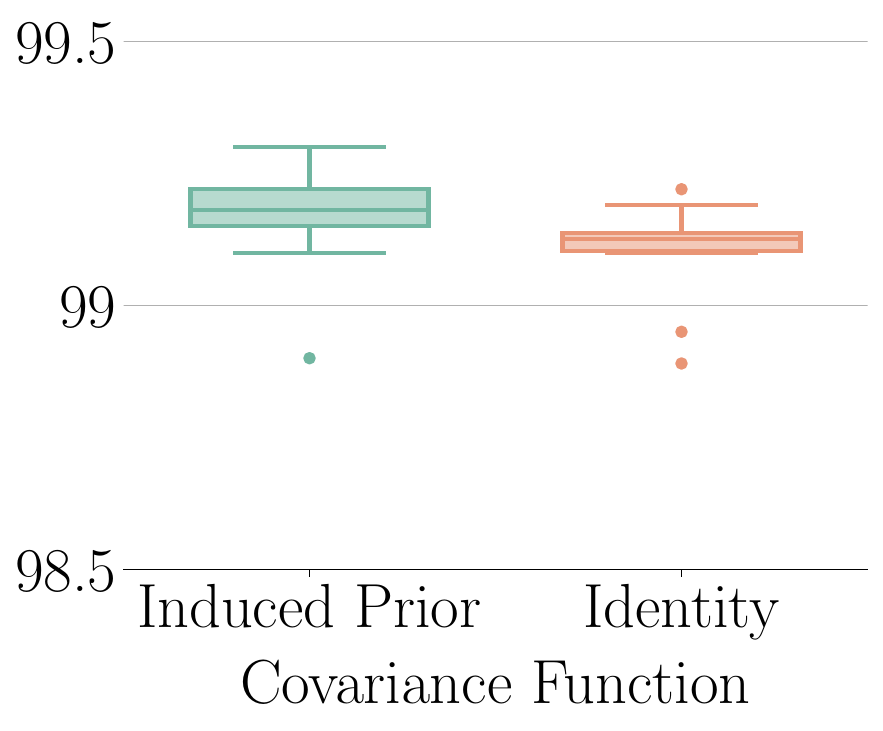}
    }
    \subfloat[\pmnist (SH)]{
        \label{fig:covariance_pmnist}
        \includegraphics[height=3.2cm, keepaspectratio]{
            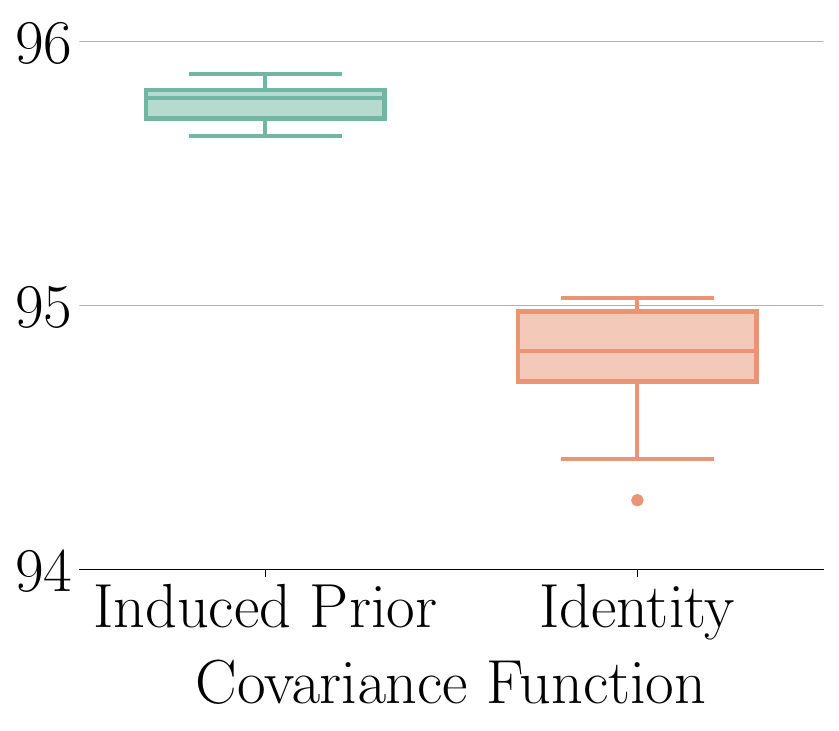}
    }
    \subfloat[\smnist (SH)]{
        \label{fig:covariance_smnist_sh}
        \includegraphics[height=3.2cm, keepaspectratio]{
            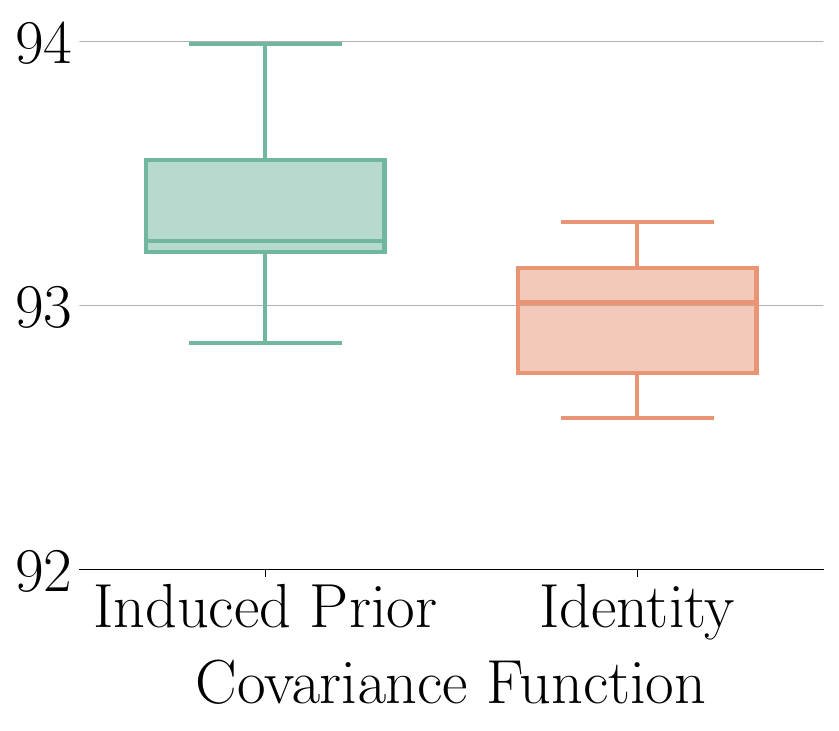}
    }
    \caption{
        \textbf{Effect of Empirical Prior Covariance.}$~$
        Comparison of predictive performance under the induced prior covariance function \mbox{$\mathbf{K}^{p_{t}} = \textrm{{diag}}\left( \jac_{\bmu_{t-1}}(\bx) \bSigma_{t-1} \jac_{\bmu_{t-1}}(\bx')^\top \right)$} (left) vs. an identity covariance function (right).
    }
    \label{fig:covariance}
    \vspace*{-10pt}
\end{figure*}

\begin{figure*}[h!]
    \subfloat[\smnist (MH)]{
        \includegraphics[height=3cm, keepaspectratio]{
            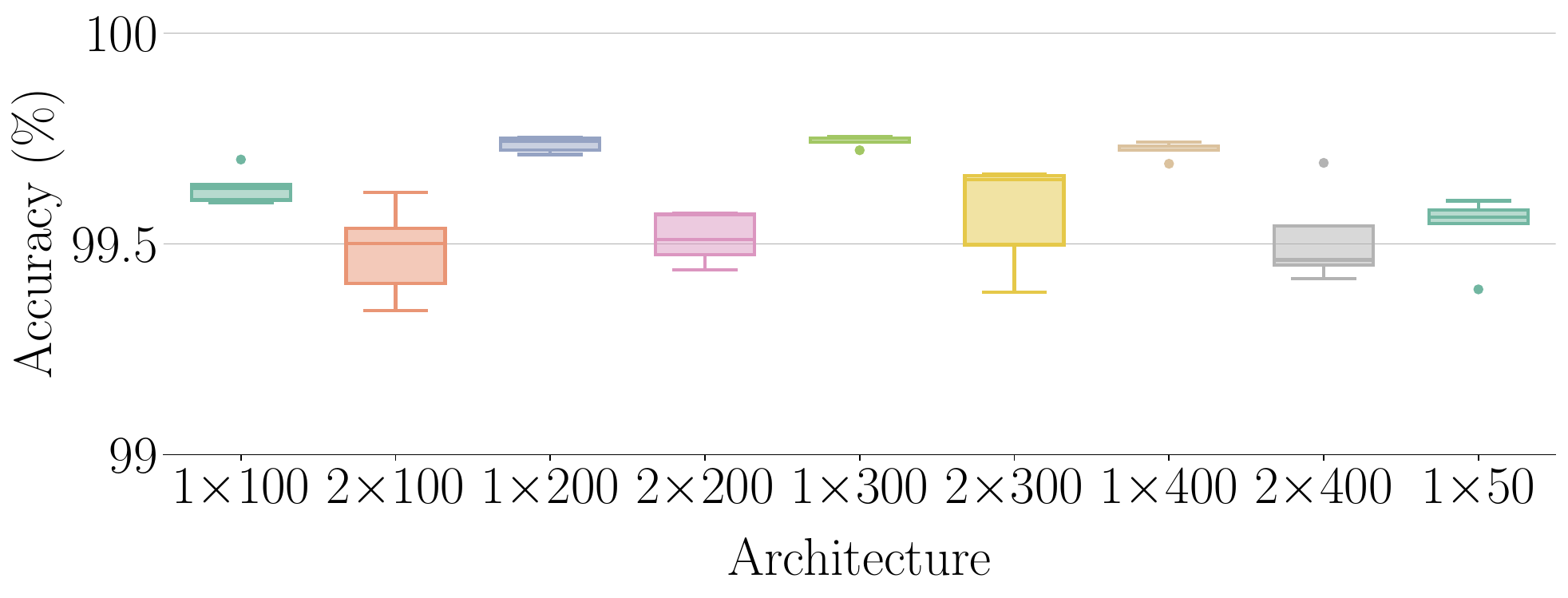}
        \includegraphics[height=3cm, keepaspectratio]{
            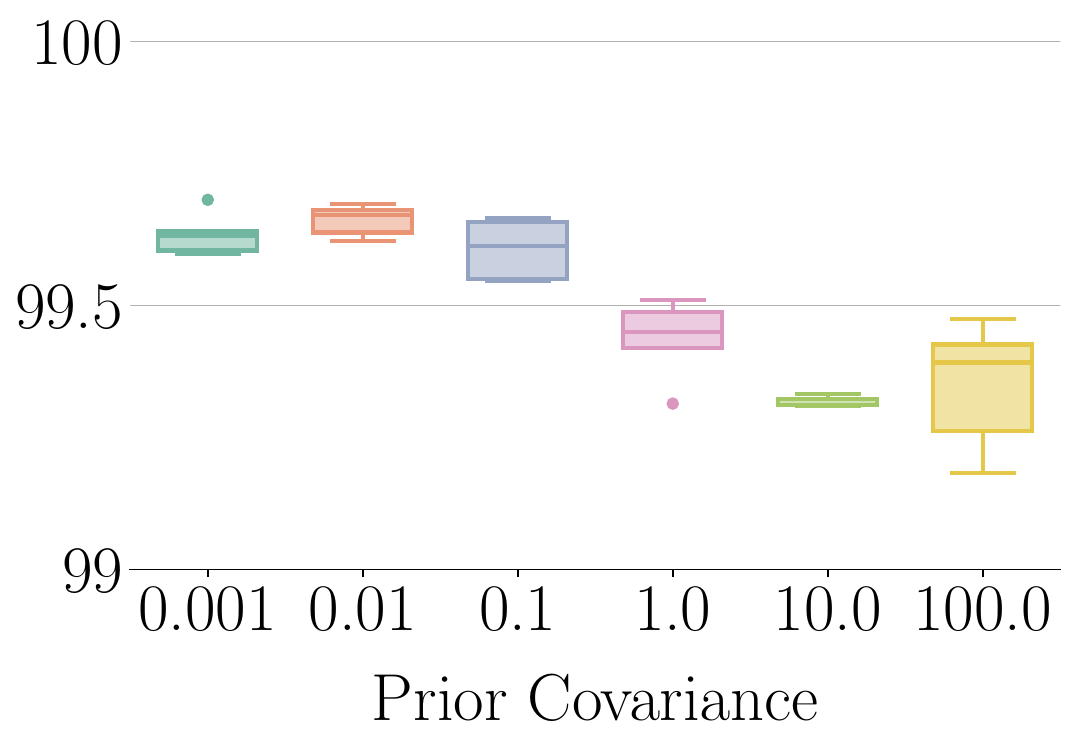}
        \includegraphics[height=3cm, keepaspectratio]{
            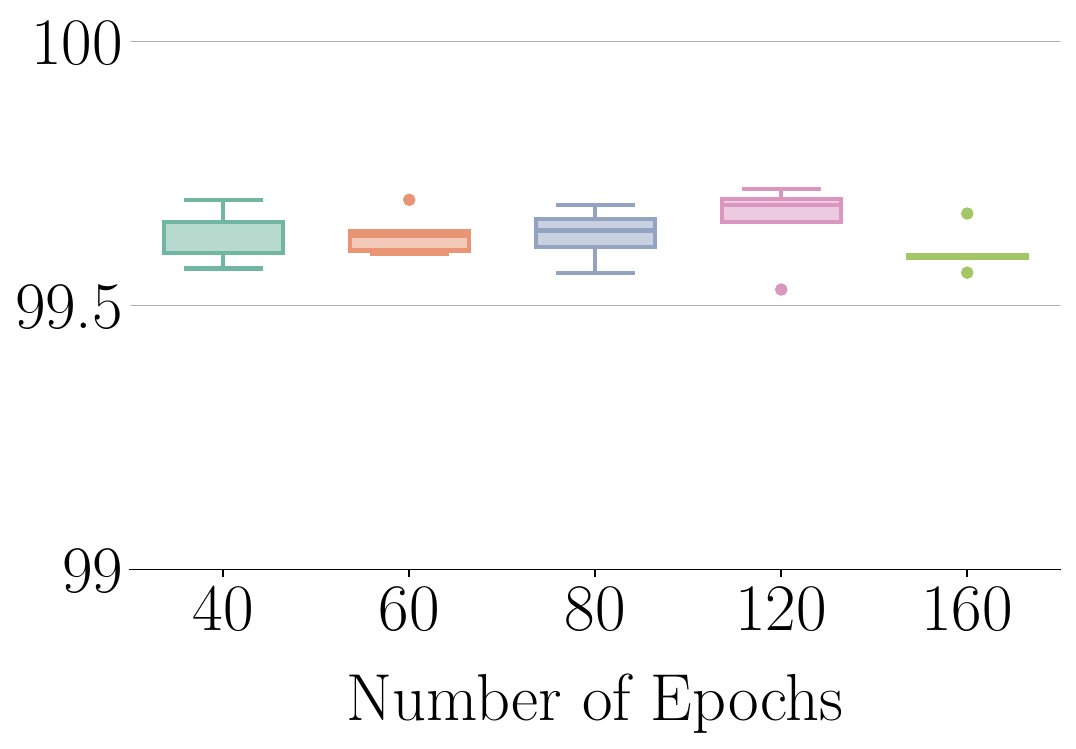}
    }
    \\
    \subfloat[\sfmnist (MH)]{
        \includegraphics[height=3cm, keepaspectratio]{
            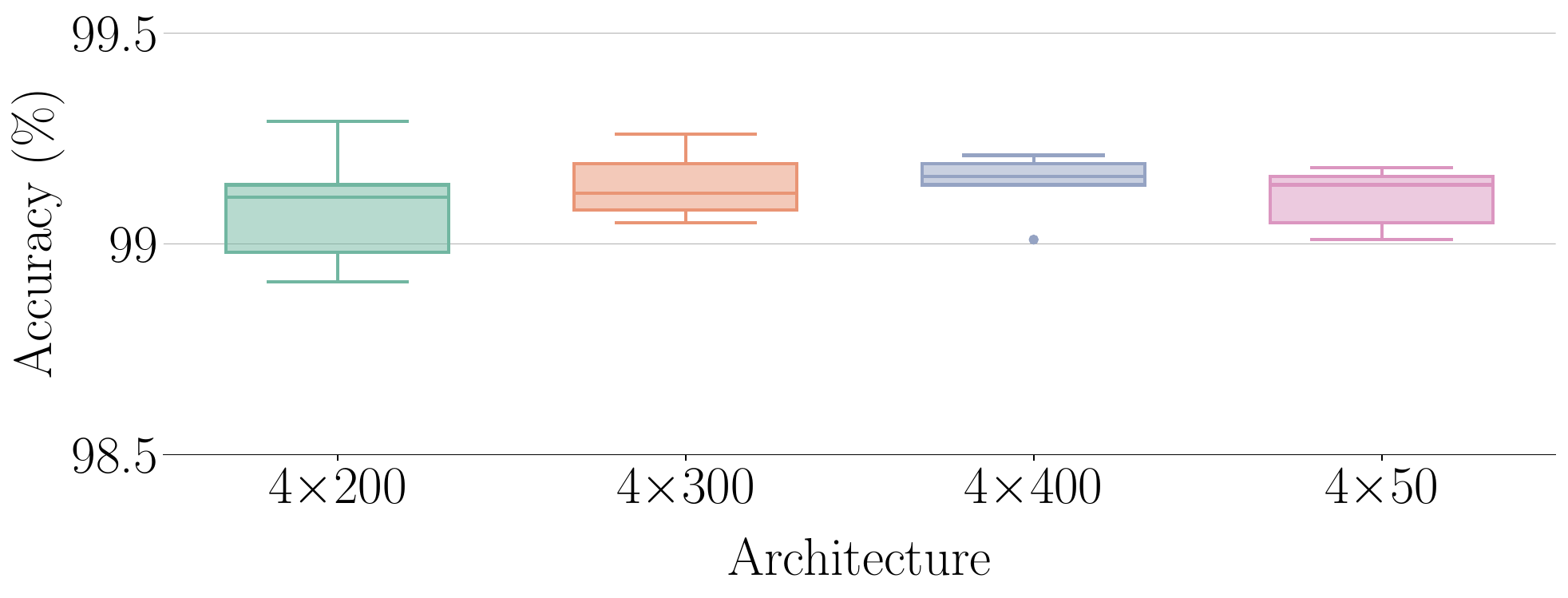}
        \includegraphics[height=3cm, keepaspectratio]{
            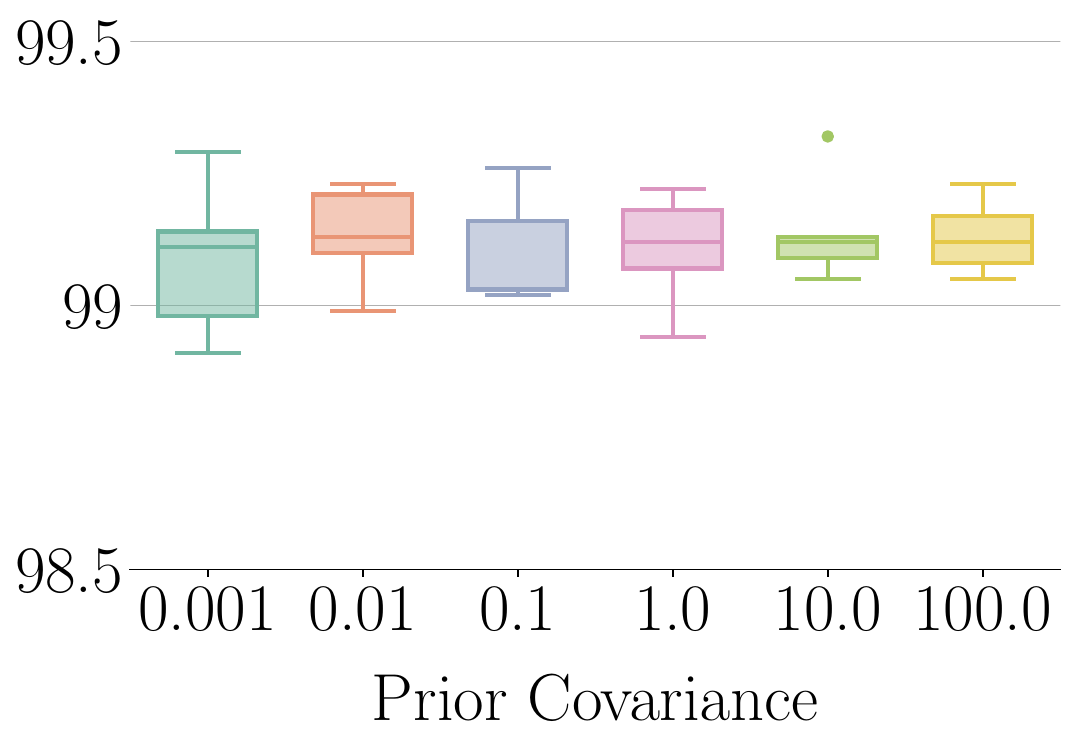}
        \includegraphics[height=3cm, keepaspectratio]{
            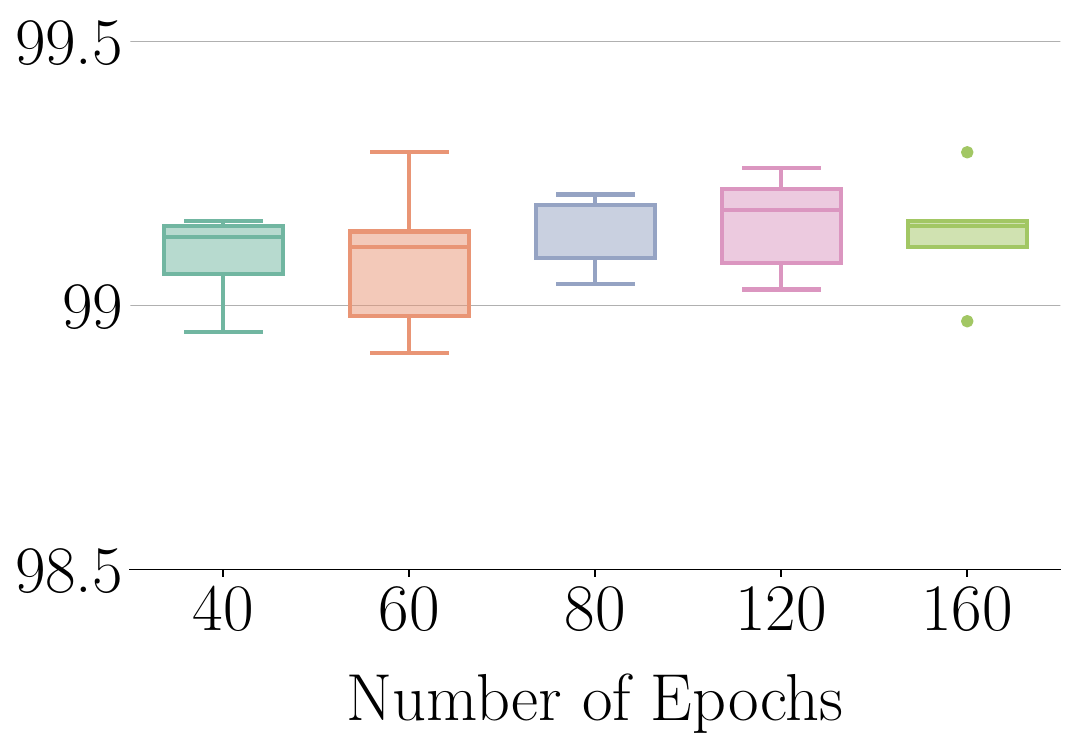}
    }
    \\
    \subfloat[\pmnist (SH)]{
        \includegraphics[height=3cm, keepaspectratio]{
            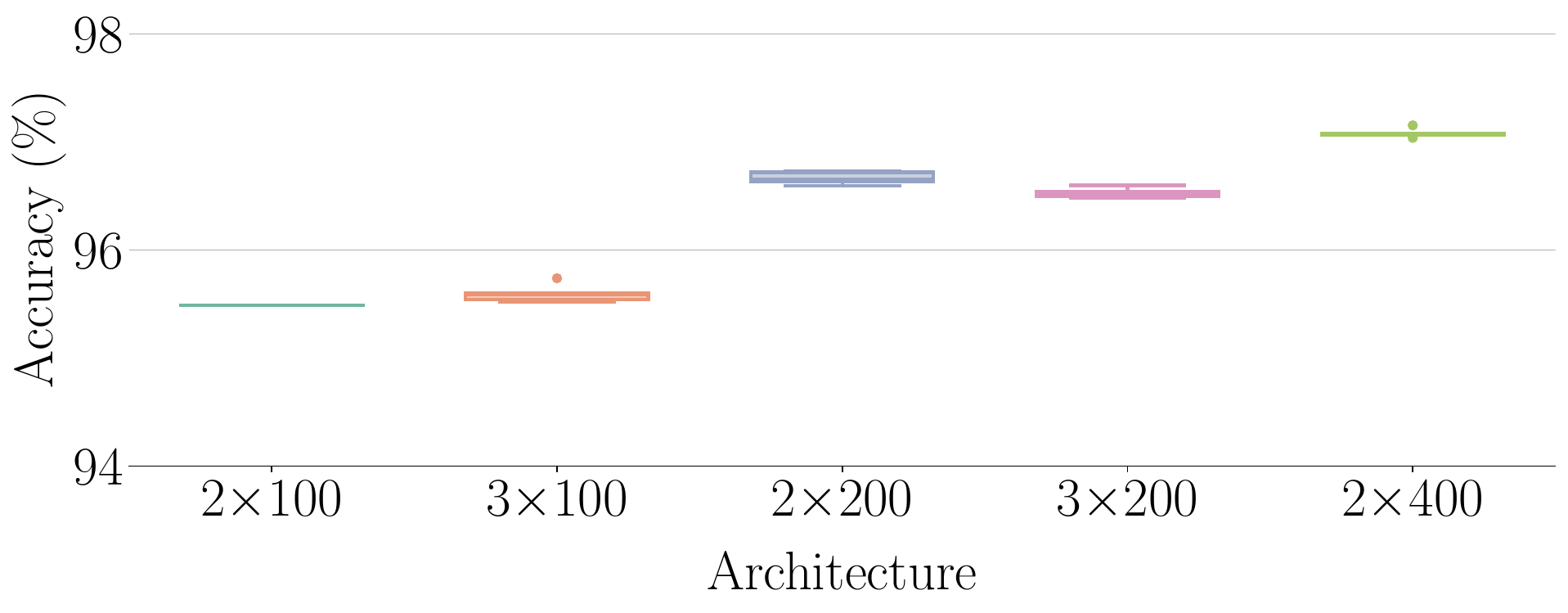}
        \hspace*{5pt}
        \includegraphics[height=3cm, keepaspectratio]{
            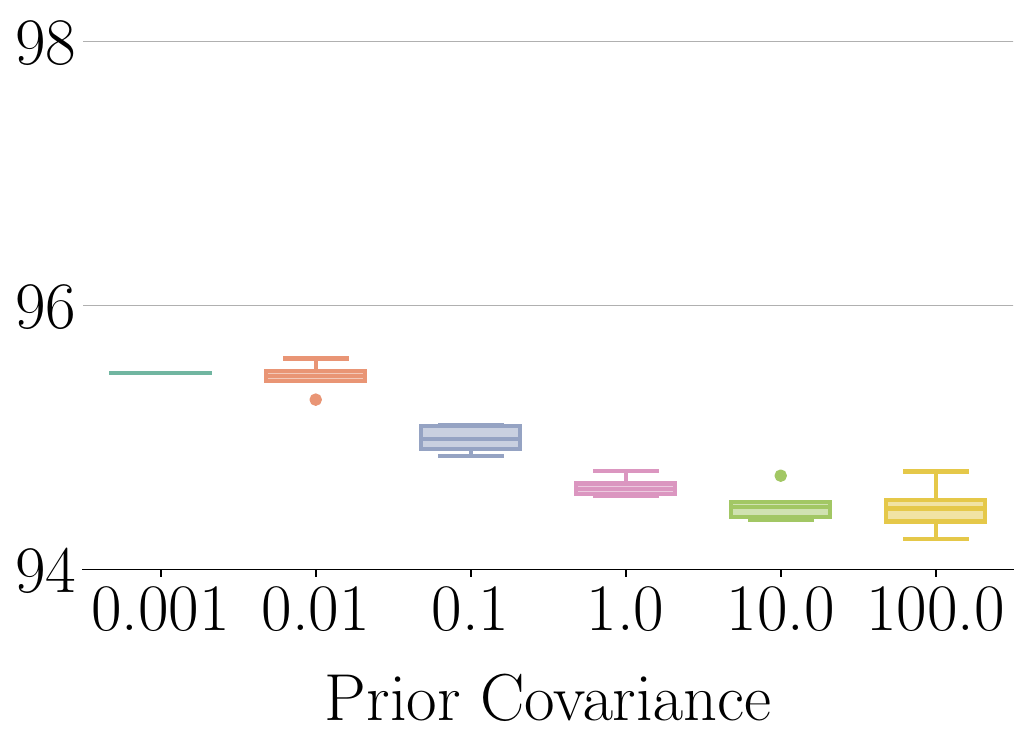}
        \hspace*{5pt}
        \includegraphics[height=3cm, keepaspectratio]{
            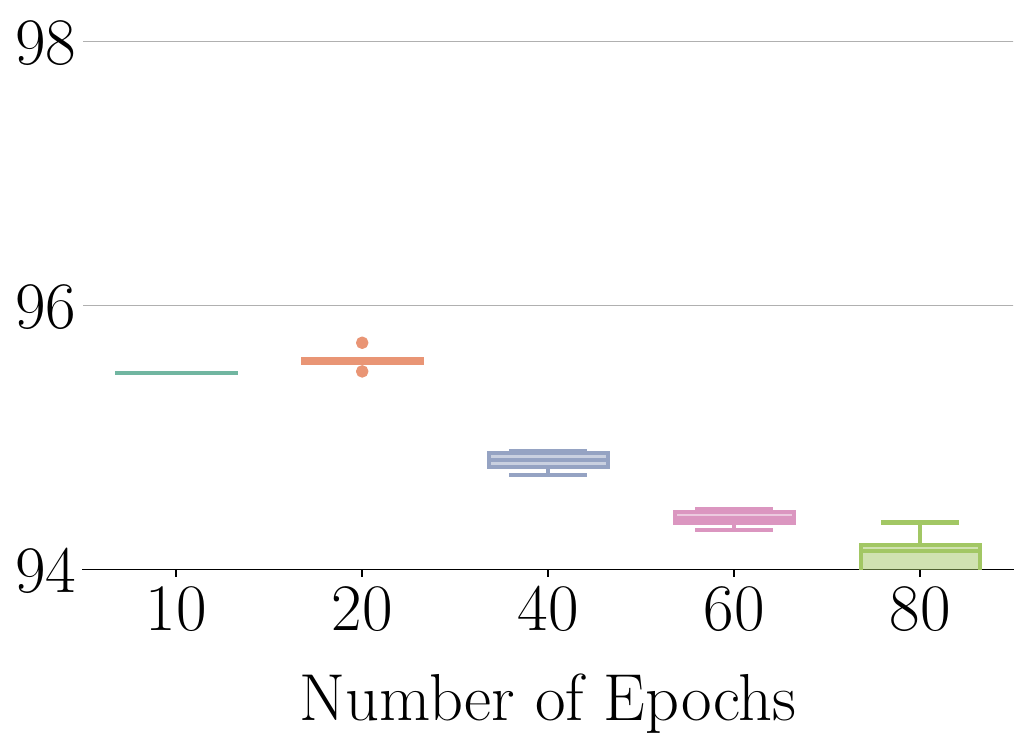}
    }
    \\
    \subfloat[\smnist (SH)]{
        \includegraphics[height=3cm, keepaspectratio]{
            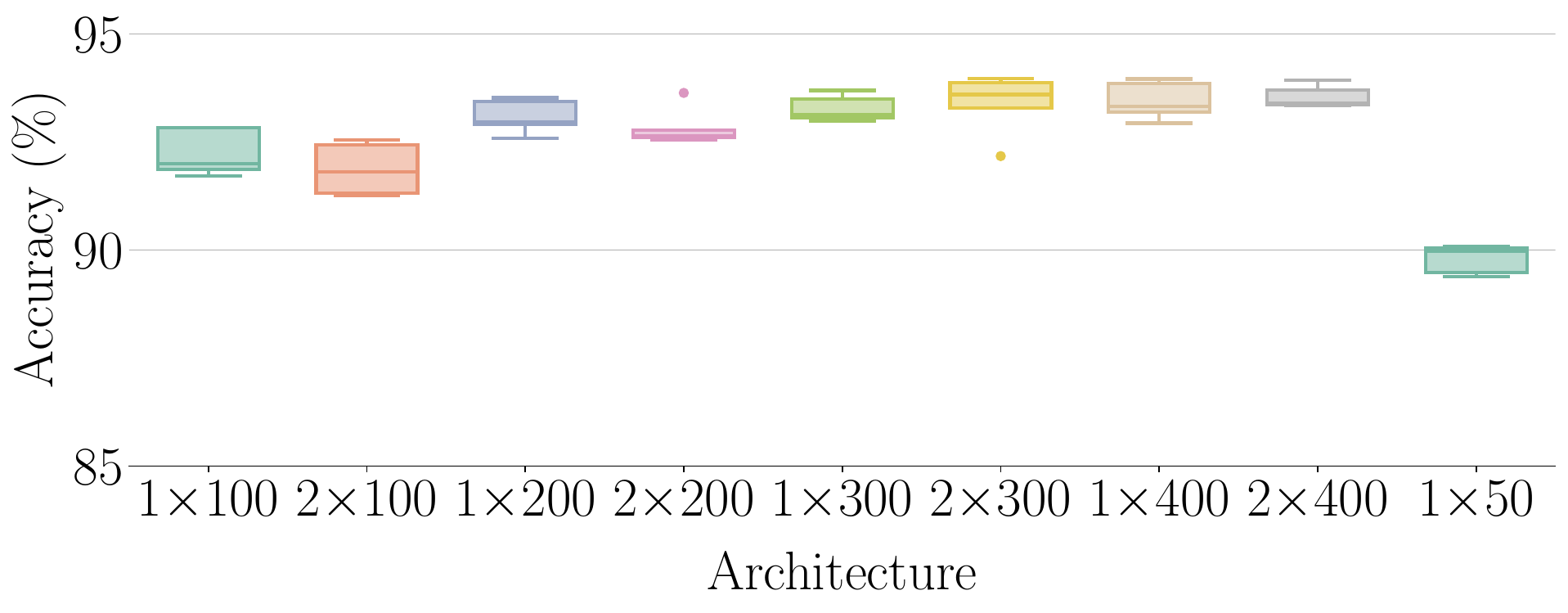}
        \hspace*{5pt}
        \includegraphics[height=3cm, keepaspectratio]{
            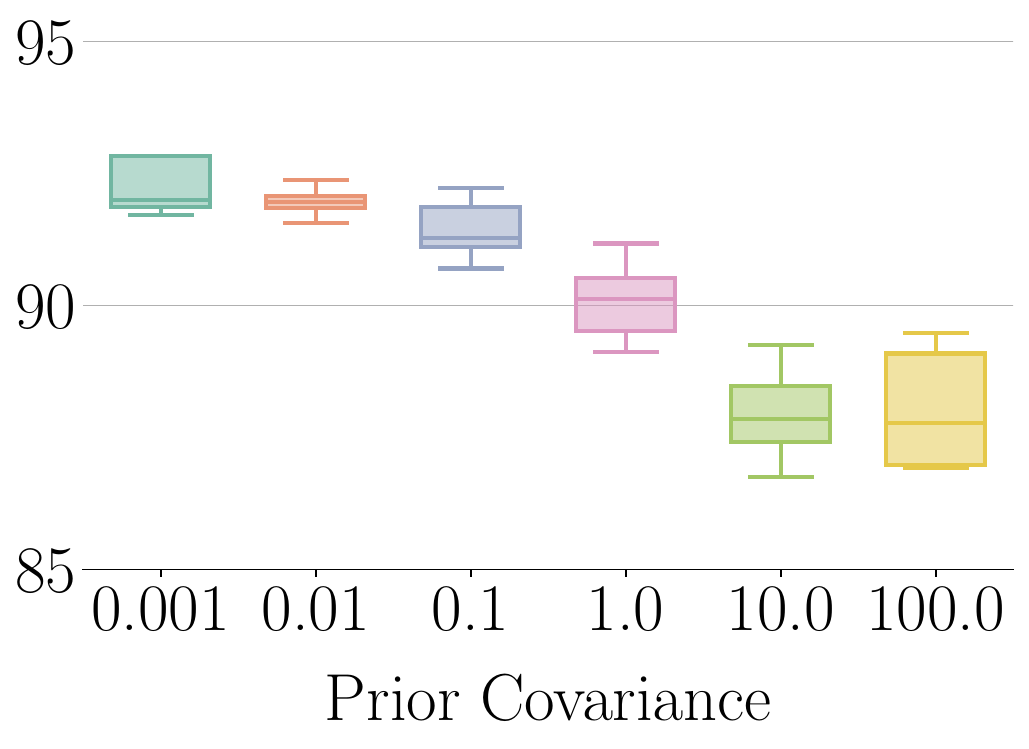}
        \hspace*{5pt}
        \includegraphics[height=3cm, keepaspectratio]{
            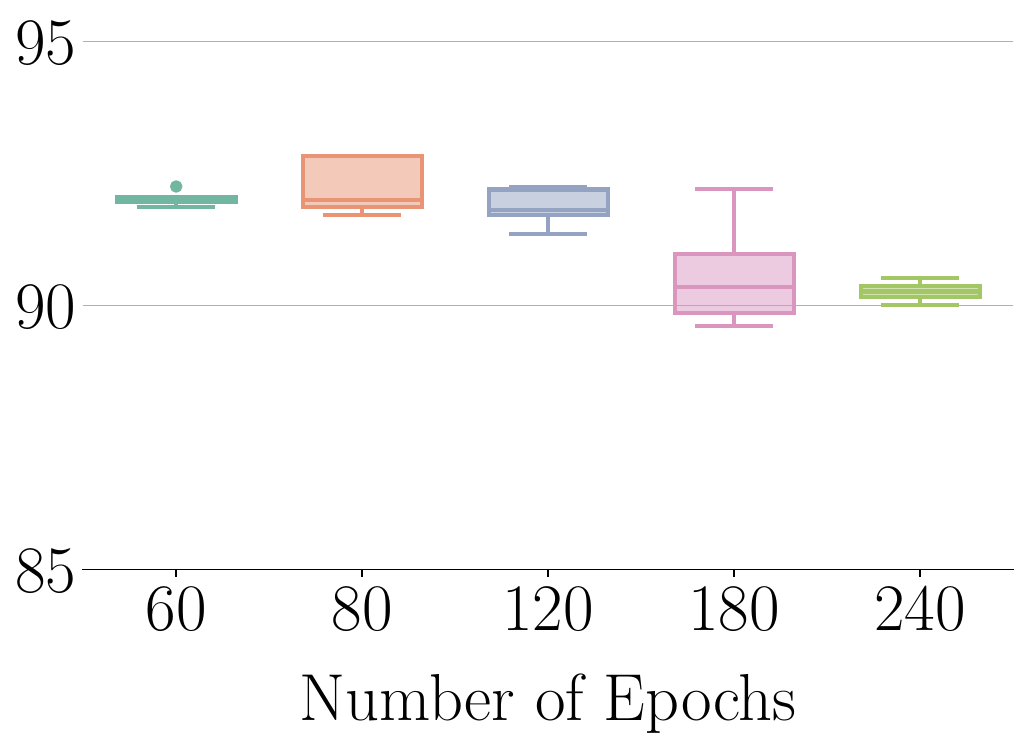}
    }
    \caption{
        \textbf{Effect of Neural-Network Size, First-Task Prior Covariance, and the Number of Training Epochs.}$~$
        We explore settings of neural-network size (e.g., $2\times100$ means a fully connected neural network with two hidden layers of size 100), initial prior covariance and number of training epochs for each task.
        To limit the computational resources required, we vary the values of one hyperparameter at a time instead of carrying out a full grid search.
    }
    \label{fig:hyper_mnist}
    \vspace*{-50pt}
\end{figure*}

\clearpage

\begin{figure*}[ht!]
    \centering
    \hspace{-10pt}
    \subfloat[Permuted MNIST (SH)]{
        \label{fig:coreset_minimal_pmnst}
        \includegraphics[height=3.2cm, keepaspectratio]{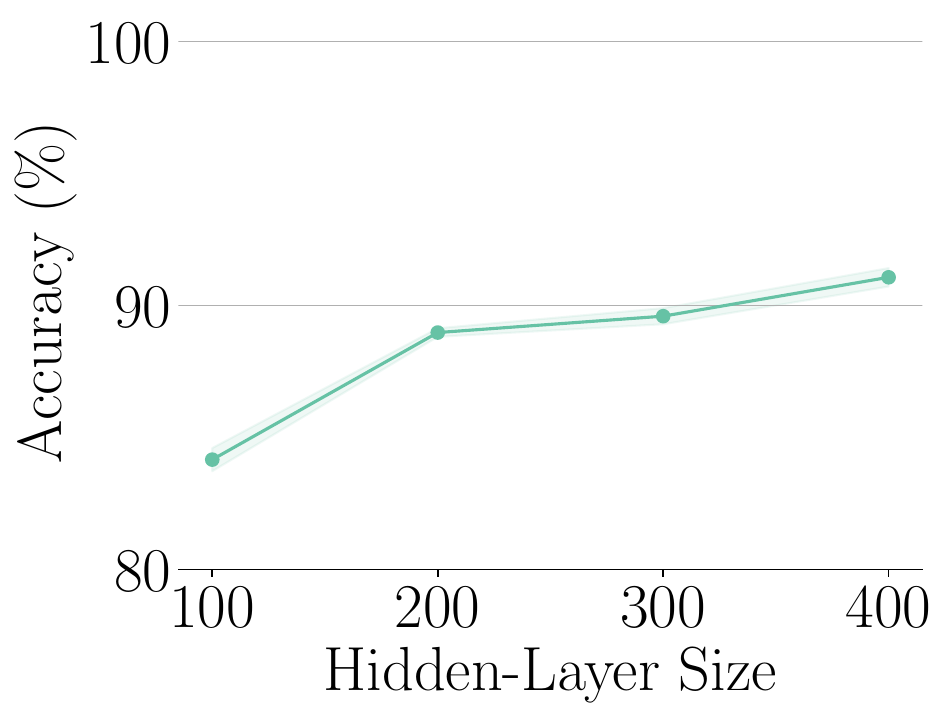}
    }
    \hspace{0.1\textwidth}
    \subfloat[Split MNIST (SH)]{
        \label{fig:coreset_minimal_smnst_sh}
        \includegraphics[height=3.2cm, keepaspectratio]{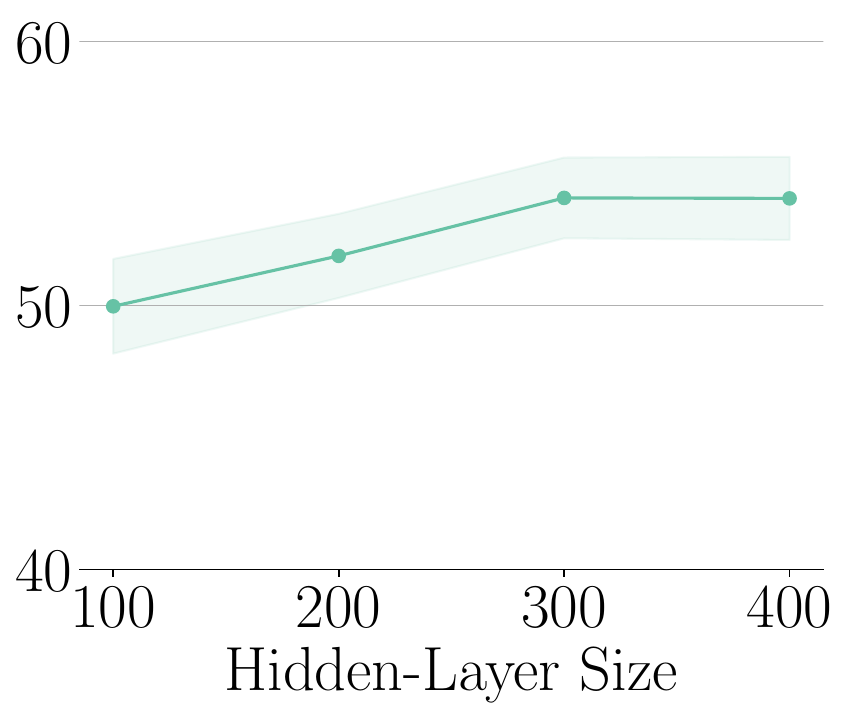}
    }
    \caption{
        \textbf{Effect of Neural-Network Size under Minimal Coresets.}$~$
        Predictive accuracy under \sfsvi on permuted \mnist (SH) and split \mnist (SH)  as a function of network width, using only a minimal coreset of one sample per class, selected randomly.
    }
    \label{fig:coreset_minimal}
\end{figure*}

\begin{figure*}[ht!]
    \centering
    \subfloat{
        \includegraphics[height=3.2cm]{
            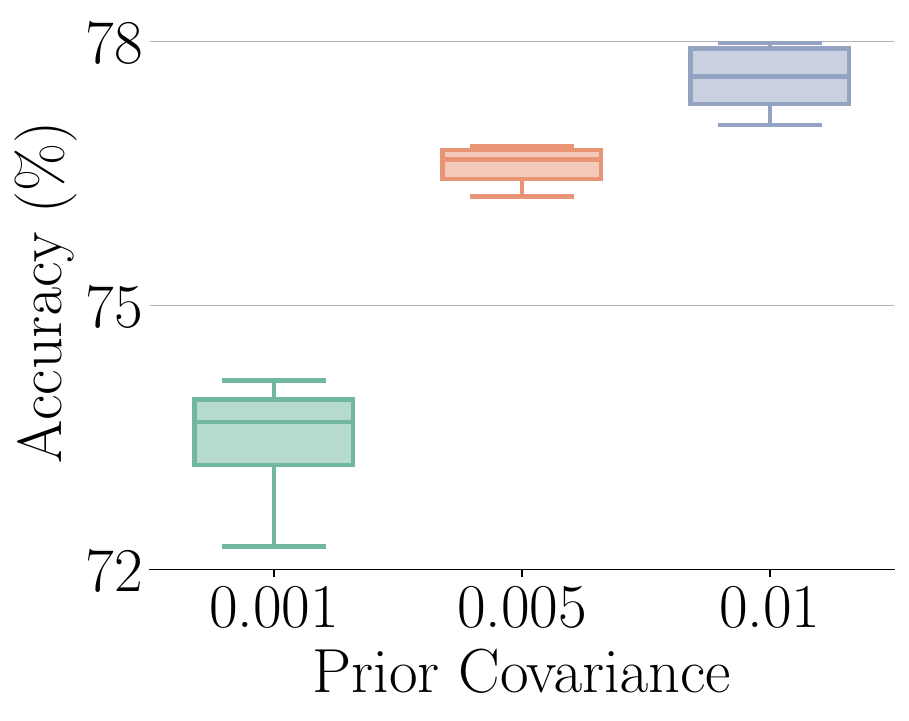}
    }
    \hspace{0.1\textwidth}
    \subfloat{
        \includegraphics[height=3.2cm]{
            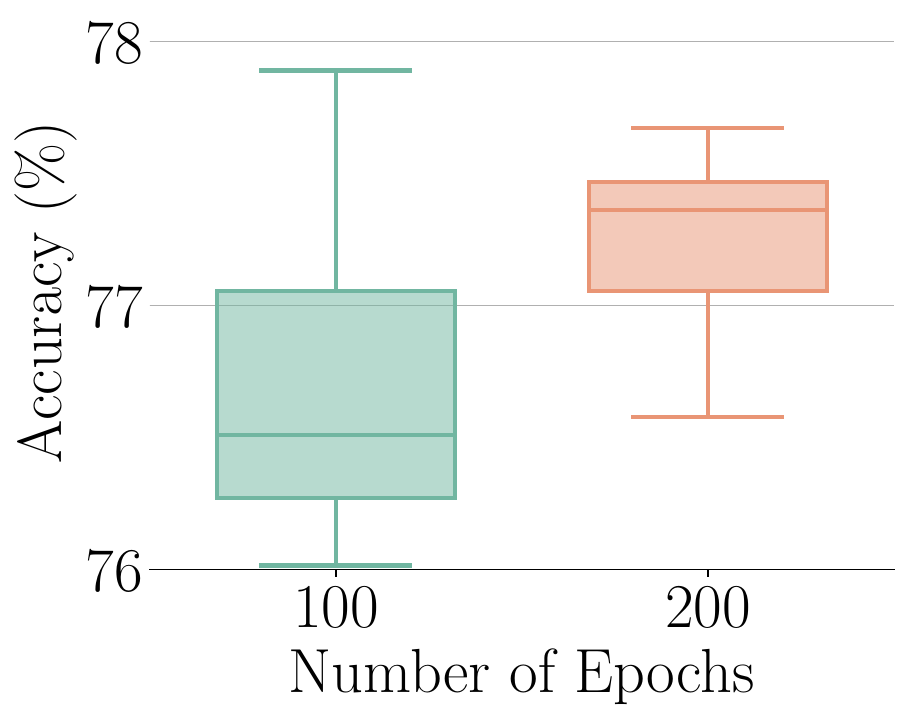}
    }
    
    \caption{
        \textbf{Hyperparameter Search on Split CIFAR.}$~$
        We explore different settings of the initial first-task prior covariance and the number of epochs for the first task.
        To limit the computational resources required, we vary the values of one hyperparameter at a time instead of carrying out a full grid search.
    }
    \label{fig:hyper_cifar}
\end{figure*}

\begin{figure*}[ht!]
    \centering
    \includegraphics[height=3.6cm, keepaspectratio]{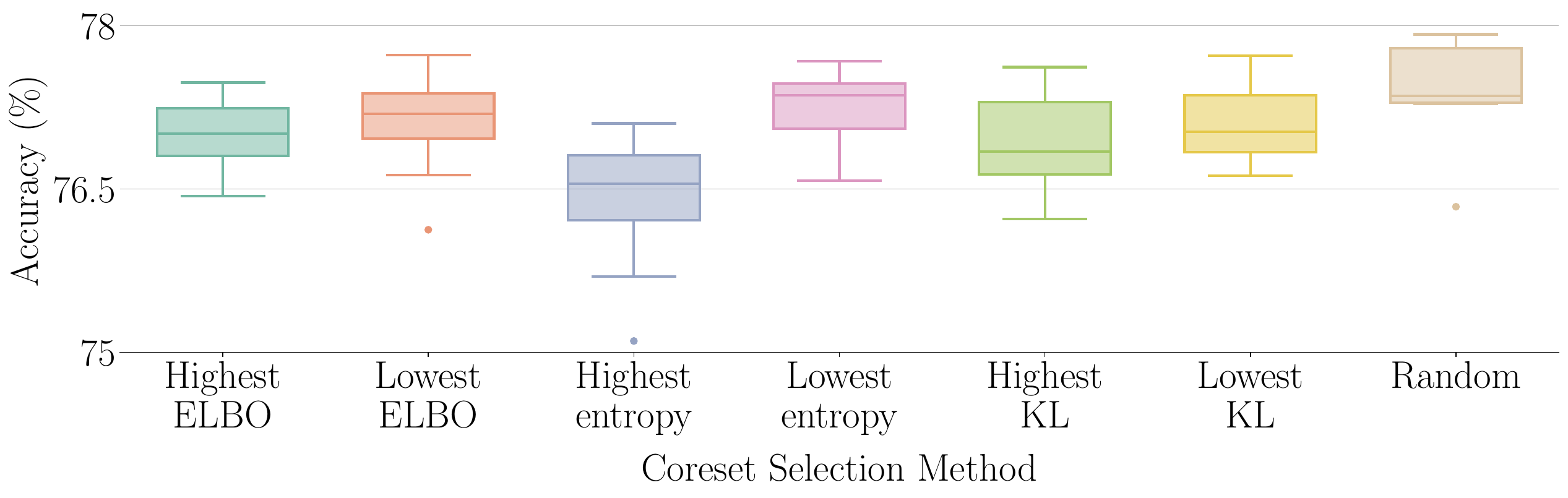}
    \caption{
        \textbf{Comparison of Different Coreset-Selection Methods on Split CIFAR.}$~$
        For score-based coreset-selection methods, we first score each coreset point---using~\Cref{eq:variational_objective_diag_paper} for \elbo scoring, using the predictive entropy for entropy scoring, and the KL divergence in~\Cref{eq:variational_objective_diag_paper} for KL scoring---then sample context points from the coreset according to the probability mass function defined in~\Cref{eq:pmf_score}.
    }
    \label{fig:hyper_cifar_coreset}
\end{figure*}

\begin{figure*}[ht!]
    \centering
    \includegraphics[height=3.2cm, keepaspectratio]{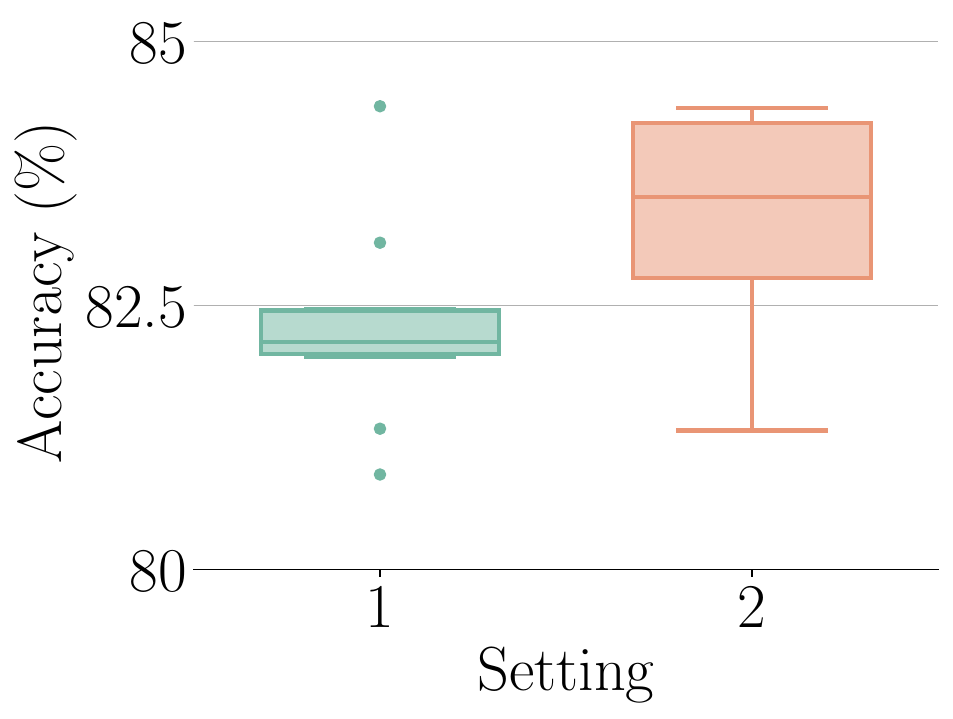}
    \caption{
        \textbf{Hyperparameter Search on Sequential Omniglot.}$~$
        We compare two settings.
        In the first, we always sample one context point for each previous task from the context set at each gradient step.
        In the second, we sample a larger number of context points (with a budget of 60 samples per gradient step) from the context set when learning on the first 25 tasks.
    }
    \label{fig:hyper_omniglot}
    \vspace*{-20pt}
\end{figure*}

\clearpage

\section{Experimental Details}
\label{appsec:experiment_details}

Our empirical evaluation centers around six sequences of classification tasks: a synthetic sequence of binary-classification tasks with 2D inputs; split \mnist; split Fashion \mnist; permuted \mnist; split \cifar; and sequential Omniglot.
With the exception of permuted \mnist, each of these task sequences can be tackled by a neural network with either a multi-head setup (MH) or a single-head setup (SH).
In a multi-head setup, the neural network has a separate output layer (or head) for each task, and task identifiers are provided at test time in order to select the appropriate head. 
In a single-head setup, the neural network has just one output layer shared across all tasks, and task identifiers are not provided.
In our experiments, we use multi-head setups for split Fashion \mnist, split \cifar and sequential Omniglot, and single-head setups for the synthetic task sequence along with permuted \mnist.
For split \mnist, we run both setups.

\subsection{Illustrative Example}

The task sequence shown in~\Cref{fig:2d_toy} was created by~\citet{Pan2020ContinualDL}.
Each of the five tasks in this sequence involves binary classification on 2D inputs, where the number of training examples per task is 3,600.
Following~\citet{Pan2020ContinualDL}, we use a fully connected neural network with an input layer of size 2, two hidden layers of size 20 and an output layer of size 2.
When running \sfsvi, we set the prior covariance as $\bSigma_{0}=0.1$ and train the neural network for 250 epochs on each task.
We use the Adam optimizer with an initial learning rate of $0.0005$ ($\beta_1=0.9, \beta_2=0.999$) and a batch size of 128.
The coreset is constructed by choosing 40 samples from the training data for each task.
To evaluate the \kld between the posterior and the prior distributions over functions, for each previous task we sample 20 input points from the context set and generate another 30 samples by sampling each pixel uniformly from the range $[-4, 4]$.
For example, when we train the model on task $t \in \{1, 2, 3, \ldots\}$, we use $20 (t - 1)$ samples chosen from the context set and $30 t$ white-noise samples.
The noise samples encourage the neural network to preserve high predictive uncertainty in regions far from the training data.

\subsection{Task Sequences Based on (Fashion) MNIST}
\label{appsec:training_details}

Split \mnist consists of five tasks, where each task is binary classification on a pair of \mnist classes.
Split Fashion \mnist has the same form but uses data from Fashion \mnist.
Permuted \mnist comprises ten tasks, where each task involves classifying images into the ten \mnist classes after the image pixels have been randomly reordered.
Unless specified otherwise, the following setups apply to~\Cref{fig:coreset_heuristic,fig:coreset_none,fig:vcl,fig:covariance} and~\Cref{tab:results}.

\textbf{Dataset.}$~$
In all cases, 60,000 data samples are used for training and 10,000 data samples are used for testing.
The input images are converted to floating-point numbers with values in the range $[0, 1]$.

\textbf{Neural-Network Size \& Coreset Size.}$~$
To ensure fair comparison, all methods in~\Cref{tab:results} (unless where explicitly indicated otherwise) use the same neural-network size and (where applicable) coreset size.
As in prior work~\citep{Pan2020ContinualDL,titsias2020functional}, we use fully connected neural networks, with two hidden layers of size 100 for permuted \mnist and two hidden layers of size 256 for split (Fashion) \mnist.
In all cases, the ReLU activation function is applied to non-output units.
For single-head setups, we use 200 coreset points; for multi-head setups, we use 40 points.

\textbf{Coreset Selection.}$~$
For \sfsvi with a coreset, when training on the first task, 40 context points are generated by sampling each pixel uniformly from the range $[0, 1]$; during training on subsequent tasks, 40 context points are chosen randomly from the context set.
For \sfsvi without a coreset, 40 context points are chosen uniformly randomly from the training data of the current task (corresponding to the ``Random'' label in~\Cref{fig:coreset_heuristic}).

\textbf{Prior Distribution.}$~$
For the first task, \sfsvi uses a prior distribution over functions with fixed mean and diagonal covariance.
When using a coreset, the prior distribution is assumed to be Gaussian with zero mean and a diagonal covariance of magnitude 0.001.
When not using a coreset, the prior distribution is assumed to be Gaussian with zero mean and a diagonal covariance of magnitude 100.
The prior variance is optimized via hyperparameter selection on a validation set.

\textbf{Optimization.}$~$
We use the Adam optimizer with an initial learning rate of $0.0005$ ($\beta_1=0.9, \beta_2=0.999$).
The number of epochs on each task is 60 for split \mnist (MH), 60 for split Fashion \mnist (MH), 10 for permuted \mnist (SH) and 80 for split \mnist (SH).
The batch size is 128. 

\textbf{Prediction.}$~$
The predictive distribution used for computing the expected log-likelihood is estimated using five Monte Carlo samples.

\textbf{Hyperparameter Selection.}$~$
For ``\sfsvi (optimized)'' in~\Cref{tab:results}, we used the optimized hyperparameters chosen on a validation set after exploring the configurations shown in~\Cref{tab:hyperparameter}.
For cases where no configuration is significantly better than the rest, the default value given in~\Cref{appsec:training_details} is used.

\begin{table*}[t!]
    \caption{
        Hyperparameter selection.
        Optimal values (in bold) were chosen based on validation-set accuracy.
        Standard errors were computed across ten random seeds.
    }
    \centering
    \small
    \vspace*{2pt}
    \label{tab:hyperparameter}
    \resizebox{\columnwidth}{!}{%
    \begin{tabular}{l c c c}
        \toprule
        Task Sequences & Number of Layers \& Units & Magnitude of Prior Variance & Number of Epochs \\
        \midrule
        Split \mnist (MH) & 
            \{\textbf{1}, 2\} * \{100, 200, 300, \textbf{400}\} & 
            \{\textbf{0.001}, 0.01, 0.1, 1, 10, 100\} & 
            \{40, \textbf{60}, 80, 120, 160\} \\
        Split Fashion \mnist (MH) & 
            $~~$\{\textbf{4}\} * \{50, \textbf{200}, 300, 400\} & 
            \{\textbf{0.001}, 0.01, 0.1, 1, 10, 100\} & 
            \{40, \textbf{60}, 80, 120, 160\} \\
        Permuted \mnist (SH) & 
            \{\textbf{2}\} * \{100, 200, 400, \textbf{500}\}$~~~~$ & 
            \{\textbf{0.001}, 0.01, 0.1, 1, 10, 100\} & 
            \{10, \textbf{20}, 40, 60, 80\} \\
        Split \mnist (SH) & 
            \{\textbf{1}, 2\} * \{100, 200, 300, \textbf{400}\} & 
            \{\textbf{0.001}, 0.01, 0.1, 1, 10, 100\} & 
            \{60, \textbf{80}, 120, 160, 240\} \\
        \bottomrule
    \end{tabular}
    }
    \vspace*{-8pt}
\end{table*}

\subsection{Split CIFAR}
\label{app_subsec:cifar}

Split \cifar, as described in~\citet{Pan2020ContinualDL}, consists of six tasks.
The first is ten-way classification on the full \cifar-10 dataset.
Each of the following five is also ten-way classification, with classes drawn from \cifar-100.
Following~\citet{Pan2020ContinualDL}, we use a neural network with four convolutional layers followed by two fully connected layers followed by multiple output heads (one for each task).
For \sfsvi, we use the following setup: Adam optimizer with learning rate 0.0005, prior with covariance 0.01, random coreset selection, 200 coreset points per task, 50 context points at each task.
We also use this setup (and a training duration of 2000 epochs) when training individual neural networks for the ``separate'' baseline.

\subsection{Sequential Omniglot}
\label{app_subsec:omniglot}
Sequential Omniglot, as described in~\citet{schwarz2018progress}, comprises 50 classification tasks.
Each task is associated with an alphabet, and the number of characters (classes) varies between alphabets.
Following \citet{schwarz2018progress}, we use a neural network with four convolutional layers followed by one fully connected layer.
For \sfsvi, we use two coreset points per character, as used by~\citet{titsias2020functional}.
The coreset points are sampled from the training set with probability proportional to the entropy of the neural network's posterior predictive distribution.
To limit memory usage, we draw no more than 25 context points from the context set at each gradient step after task 25.
We use a learning rate of 0.001 and a prior covariance of 1.0.
For the first task, the neural network trains for 200 epochs; for subsequent tasks, it trains for ten epochs per task.
We use the same data augmentation and train-test split as~\citet{titsias2020functional}.

\subsection{Coreset-Selection Methods}

We consider different distributions from which to sample points to be added to the coreset.
For each of the scoring methods below, we use the scores to create a probability mass function from which points can be sampled.

\textbf{Random.}$~$
Points are sampled uniformly from the training data.

\textbf{Predictive-Entropy Scoring.}$~$
Points are scored according to the total predictive uncertainty (i.e., the predictive entropy) of the model.
For a model with stochastic parameters $\bTheta$, pre-likelihood outputs $f(\bX ; \btheta)$, and a likelihood function $p(\by \vbar f(\bX ; \btheta))$, the predictive entropy is given by $\mathcal{H}(\E [ p(\by \vbar f(\bX ; \btheta)) ] )$~\citep{cover1991elements,shannon1949mathematical}.
The expectation is taken with respect to the model parameters.
$\mathcal{H}(\cdot)$ is the entropy functional, and $\mathcal{I}(\by_\ast ;\, \bTheta)$ is the mutual information between the model parameters and its predictions.

\textbf{Evidence-Lower-Bound Scoring.}$~$
Points are scored according to the value of the evidence lower bound (ELBO) given in~\Cref{eq:variational_objective_diag_paper}.

\textbf{Kullback-Leibler-Divergence Scoring.}$~$
Points are scored according to the value of the approximation to the function-space \kld given in~\Cref{eq:variational_objective_diag_paper}.

\textbf{Score-Based Distributions.}$~$
After scoring with the above methods, points are added to the coreset by sampling from one of the following probability mass functions:
\begin{align}
\label{eq:pmf_score}
\begin{split}
    \textrm{\textbf{Lowest:}}\quad \mathbb{P}(i) \defines \frac{\bar{s}_i}{\sum_{j=1}^N \bar{s}_j}
    \qquad \textrm{and} \qquad
    \textrm{\textbf{Highest:}}\quad \mathbb{P}(i) \defines \frac{s_i}{\sum_{j=1}^N s_j} ,
\end{split}
\end{align}
where $s_i$ is the score of $i$-th point, $\bar{s}_i = \max_{j=1}^N s_j - s_i$, and $N$ is the number of candidate points.

\subsection{Forward and Backward Transfer}
\label{appsec:forward_backward_transfer}
In~\Cref{tab:transfer}, we report forward and backward transfer metrics as defined in~\citet{Pan2020ContinualDL}.
Backward transfer (BT) indicates the performance gain on past tasks when new tasks are learnt, while forward transfer (FT) quantifies how much knowledge from past tasks helps the learning of new tasks.
Higher is better for both.
For $T$ tasks, let $R_{i, i}$ be the accuracy of model on task $t_i$ after training on task $t_i$, and let $R_{i}^{\textrm{ind}}$ be the accuracy of an independent model trained only on task $t_i$.
Then
\begin{align*}
\begin{split}
    \textrm{BT} \defines \frac{1}{T-1} \sum_{i=1}^{T-1} R_{T, i} - R_{i, i}
    \qquad \textrm{and} \qquad
    \textrm{FT} \defines \frac{1}{T-1} \sum_{i=2}^{T} R_{i, i} - R_{i}^{\textrm{ind}} .
\end{split}
\end{align*}%

\section{Further Related Work}
\label{appsec:related_work}

Objective-based approaches to continual learning involve training a neural network using a specially designed objective function.
Typically the objective includes a regularization term that penalizes changes in the neural network's configuration.
Whereas in~\Cref{sec:related_work} we summarise methods that regularize in function space, here we cover methods that regularize directly in terms of the parameters of a neural network.
Among these, most relevant to our work are those that approximate Bayesian updating, in which the posterior from the previous task forms the prior for the current task.

A key idea is shared between many methods for parameter-space regularization: for each parameter, apply a penalty on the difference between its current setting and its prior setting, weighted by a measure of the parameter's importance.
Methods vary in how they measure importance.
Variational continual learning (\vcl;~\citealp{Nguyen2018VariationalCL,Swaroop2019ImprovingAU}), which extends the concept of online variational inference \citep{broderick2013streaming,ghahramani2000online,honkela2003online,sato2001online} to deep neural networks, uses the parameter covariance matrix of the model currently serving as the prior.
Elastic weight consolidation (\textsc{ewc};~\citealp{Kirkpatrick2017OvercomingCF}) and its successors~\citep{Chaudhry2018RiemannianWF,Lee2017OvercomingCF,Liu2018RotateYN,schwarz2018progress} use a Fisher information matrix computed on each task.
Online structured Laplace~\citep{Ritter2018OSL} and second-order loss approximation~\citep{Yin2020SOLACL} respectively use Kronecker-factored and low-rank Hessians.
Synaptic intelligence (\textsc{si};~\citealp{zenke17}) uses a cumulative sum of the gradient of the training objective with respect to the parameters.
Memory-aware synapses (\textsc{mas};~\citealp{aljundi18}) use the gradient of the model output with respect to the parameters.

Other related work on parameter-space regularization includes various modifications to \vcl ~\citep{ahn19,kessler2019hierarchical}, uncertainty-guided continual learning in Bayesian neural networks~\citep{Ebrahimi2020UncertaintyguidedCL}, and a variation of \textsc{si} known as asymmetric loss approximation with single-side overestimation~\citep{Park2019ContinualLB}.
There have also been efforts to conceptually unify some of the approaches outlined above:~\citet{Loo2020GeneralizedVC} draws a link between \vcl and online \textsc{ewc};~\citet{Chaudhry2018RiemannianWF} combines \textsc{ewc} and \textsc{si} in a single method;~\citet{Yin2020OptimizationAG} generalizes \textsc{ewc}, online structured Laplace, \textsc{si} and \textsc{mas}.

\end{appendices}

\end{document}